\documentclass[11pt,a4paper]{article}
\pdfoutput=1
\usepackage{fullpage}
\usepackage{times,latexsym}
\usepackage{url}
\usepackage[T1]{fontenc}

\usepackage{rotating}
\usepackage{blindtext}
\usepackage{amsmath}
\usepackage{amsfonts}
\usepackage{graphicx}
\usepackage{mathtools}
\usepackage{booktabs}
\usepackage{relsize}
\usepackage{float}
\usepackage{graphics}
\usepackage{multirow}
\usepackage{lipsum}
\usepackage{placeins}
\usepackage[authoryear]{natbib}
\usepackage[unicode=true,
 bookmarks=true,bookmarksnumbered=false,bookmarksopen=false,
 breaklinks=true,pdfborder={0 0 0},pdfborderstyle={},backref=false,colorlinks=true]
 {hyperref}

\usepackage[utf8]{inputenc}
\DeclareUnicodeCharacter{01B0}{\`u}
\DeclareUnicodeCharacter{01A1}{\`o}

\usepackage{threeparttable}
\usepackage{booktabs, makecell}

\usepackage[title]{appendix}
\usepackage[cal=cm]{mathalfa}
\usepackage[colorinlistoftodos]{todonotes}
\usepackage{makecell,multirow}
\usepackage[style=base]{caption}
\usepackage[export]{adjustbox}
\usepackage[ruled,vlined]{algorithm2e}
\usepackage[subtle]{savetrees}
\DeclareMathOperator*{\KL}{KL}
\DeclareMathOperator*{\kld}{~\|~}
\DeclareMathOperator*{\gvn}{ \;\big|\;}

\DeclareMathOperator*{\E}{\mathbb{E}}

\newcommand{\vf}[1]{{\boldsymbol{#1}}}

\DeclareMathOperator*{\Dir}{Dir}
\DeclareMathOperator*{\DP}{DP}
\DeclareMathOperator*{\UDP}{UDP}
\DeclareMathOperator*{\BDP}{BDP}
\DeclareMathOperator*{\FDP}{FDP}
\DeclareMathOperator*{\BFDP}{BFDP}
\DeclareMathOperator*{\softmax}{softmax}
\renewcommand{\!}{\hspace{-0.3ex}}
\renewcommand{\;}{\hspace{0.2ex}}
\renewcommand{\,}{\hspace{0.4ex}}
\newcommand{\tabitem}{~~\llap{\textbullet}~~}

\newcommand{\remove}[1]{}
\newcommand{\mycomment}[1]{}                           
\newcommand{\warn}[1]{} 
\newcommand{\fabio}[1]{} 
\newcommand{\jamie}[1]{} 

\definecolor{purple}{RGB}{148,0,211}
\definecolor{lightblue}{RGB}{121,196,237}
\definecolor{green}{RGB}{0,158,115}
\definecolor{blue}{RGB}{143,153,233}
\definecolor{grey}{RGB}{47,79,79} 
\definecolor{gold}{RGB}{216,176,49}

\newcommand{\limited}[1]{}
\newcommand{\ignore}[1]{}

\title{
  A Variational AutoEncoder for Transformers with \\
  Nonparametric Variational Information Bottleneck
}
 \author{ 
   \textbf{James Henderson}, \\ 
   Idiap Research Institute, Switzerland \\
   \texttt{james.henderson@idiap.ch}
   \and
   \textbf{Fabio Fehr}\\
   Idiap Research Institute, Switzerland \\
   EPFL, Switzerland \\
   \texttt{fabio.fehr@idiap.ch}
   \\ }

\begin{document}
\maketitle
\interfootnotelinepenalty=10000

\begin{abstract}
  We propose a VAE for Transformers by developing a variational information bottleneck regulariser for Transformer embeddings.  We formalise the embedding space of Transformer encoders as mixture probability distributions, and use Bayesian nonparametrics to derive a nonparametric variational information bottleneck (NVIB) for such attention-based embeddings.  The variable number of mixture components supported by nonparametric methods captures the variable number of vectors supported by attention, and the exchangeability of our nonparametric distributions captures the permutation invariance of attention.  This allows NVIB to regularise the number of vectors accessible with attention, as well as the amount of information in individual vectors.  By regularising the cross-attention of a Transformer encoder-decoder with NVIB, we propose a nonparametric variational autoencoder (NVAE).  Initial experiments on training a NVAE on natural language text show that the induced embedding space has the desired properties of a VAE for Transformers.
\end{abstract}

\section{Introduction} 

Attention-based deep learning models, such as Transformers \citep{Vaswani2017,devlin-etal-2019-bert}, have achieved unprecedented empirical success in a wide range of cognitive tasks, in particular in natural language processing (NLP).  On the other hand, deep variational Bayesian approaches to representation learning, such as variational autoencoders (VAEs) \citep{kingma2014autoencoding}, have also been very influential, especially due to their variational information bottleneck (VIB) \citep{alemi2017deep,kingma2014autoencoding} for regularising the induced latent representations.  Previous VIB methods only apply to a vector space, and Transformers crucially do not use a single vector as their latent representation, instead using a set of vectors \citep{lin2020variationaltransformers, fang2021transformerbased,park-lee-2021-finetuning}.  This allows the number of vectors in a Transformer embedding to grow with the size of the input, which is essential for embedding natural language text~\citep{bahdanau2015neural}, where the size of the input can range from a single word to thousands of words.  In this paper, we propose a variational information bottleneck regulariser for set-of-vector latent representations, and use it to regularise the induced latent representation of a Transformer encoder-decoder variational autoencoder.

Variational autoencoders are encoder-decoder architectures, with VIB for regularising the amount of information which can pass through the latent representation from the encoder to the decoder.  The encoder outputs the parameters for a posterior distribution, and the decoder is trained on samples from this distribution.  Information is controlled by the noise introduced by sampling, which is regularised with the KL divergence between a prior and the posterior.  Previous work on VIB has focused almost exclusively on encoder-decoder models where the interface is a single fixed-length vector, which is not the case for Transformer encoder-decoder models.

\begin{figure}[t]
  \centering
  \begin{tabular}{c}
  \includegraphics[width=0.63\linewidth]{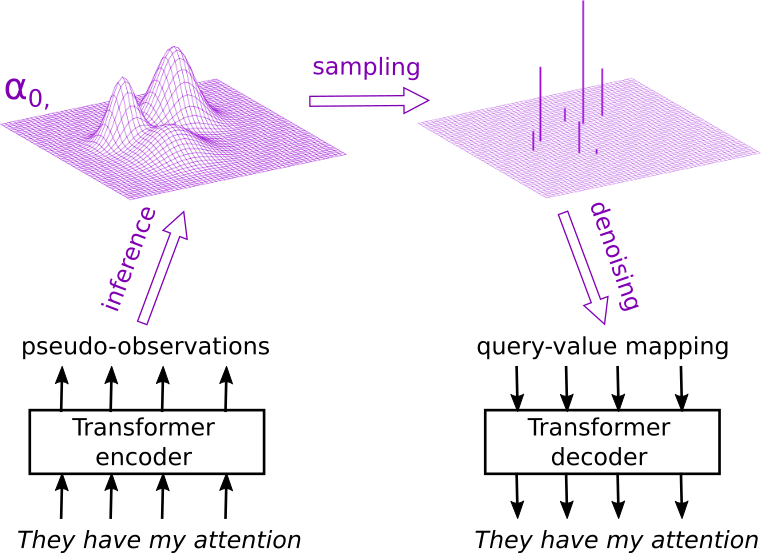}
  \\ (a) NVAE model and its NVIB layer
  \end{tabular}
  ~~~
  \begin{tabular}{c}
        \includegraphics[width=0.29\linewidth]{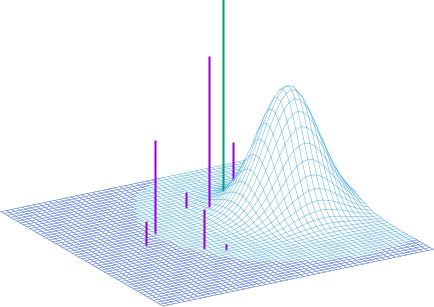}
        \\ (b) denoising attention \\[1ex]
        \includegraphics[width=0.29\linewidth]{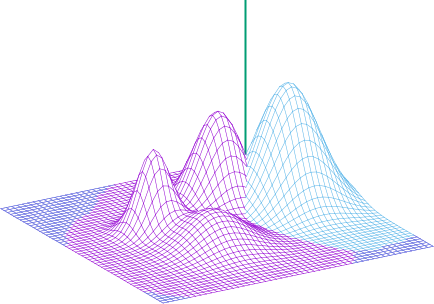}
        \\ (c) test-time denoising attention
  \end{tabular}
  \caption{(a) A depiction of the NVAE model, with its \textcolor{purple}{NVIB layer} (in purple).  A Transformer encoder embeds the sentence in the space of Dirichlet Processes, specified by a concentration parameter $\alpha_0$ plus a variable-sized exchangeable mixture distribution.  A discrete sample adds noise to this mixture distribution according to the concentration parameter.  A Transformer decoder then accesses this sample using query-denoising attention, and reconstructs the sentence.\newline
    (b) Query denoising attention at training time, with the \textcolor{purple}{sampled distribution} as the query prior, a \textcolor{lightblue}{noisy vector} as the query observation, and the \textcolor{green}{expected value} of the denoised query.\newline
    (c) \textcolor{lightblue}{Query} \textcolor{green}{denoising} attention at test time, with the \textcolor{purple}{mean distribution} as the query prior.}
  \label{fig:overview}
\end{figure}

Our approach to combining the power of Transformers with the information-theoretic regularisation of variational Bayesian methods (illustrated in Figure~\ref{fig:overview}) is based on two key properties of Transformer embeddings.  We focus on the latent representation which passes information from the encoder to the decoder in a Transformer encoder-decoder, where the decoder uses cross-attention to access the sequence of vectors output by the encoder.  Firstly, this attention function is insensitive to the order of the vectors in the sequence. This means that it interprets the output of the encoder as a permutation-invariant set of vectors \citep{deepsets2017}, not as a sequence.  Secondly, the attention function returns an interpolation between input vectors. Thus its range of output vectors does not change with the number of input vectors, and Transformer decoders generalise across sets of variable size. Therefore, Transformer encoder embeddings are interpreted by their decoder as permutation invariant and variable sized.

These properties make Transformer embeddings strikingly similar to mixture probability distributions.  Each component in a mixture distribution is a parametric distribution, thus it can be represented as a fixed-length vector of parameters.  We interpret each individual vector in a Transformer embedding as specifying a mixture component.  These components are permutation invariant, since the probability distribution does not depend on the order in which we list the mixture components - as in the permutation invariance of Transformer embeddings.  Furthermore, a mixture distribution can have any number of components - as in the variable size of Transformer embeddings.  We provide an alternative interpretation of Transformer embeddings by simply specifying one impulse distribution at each vector in the set and specifying a weighted mixture of these component distributions.  We then show that attention can be formalised as using this mixture distribution as a prior to do Bayesian query denoising (depicted in Figure~\ref{fig:overview} (b)).

This formalisation of Transformer embeddings as mixtures of impulse distributions allows us to develop a deep variational Bayesian framework for Transformer embeddings using Bayesian nonparametrics.
Bayesian nonparametrics has been widely applied to learning \textit{models} where the number of parameters grows with the size of the \textit{training data} (see \citep{Teh2010,Jordan10}), but here we apply it to inferring \textit{representations} where the number of parameters grows with the size of the \textit{input}.

Bayesian nonparametrics is a well established theory for specifying probability distributions over mixture distributions, which allows us to define prior and posterior distributions over Transformer embeddings.  In particular, we propose to use Dirichlet processes (DPs) to specify distributions over mixtures of impulse distributions.  DPs model a distribution over the effective number of components, so we can model uncertainty about the size of a Transformer embedding.    Also, DPs generate exchangeable distributions \citep{aldous1985exchangeability,Jordan10}, meaning that they capture the permutation invariance of Transformer embeddings.

This nonparametric Bayesian framework allows us to develop a variational information bottleneck for Transformer embeddings, shown in purple in Figure~\ref{fig:overview} (a).\footnote{Previously \citet{henderson-2020-unstoppable} suggested that Bayesian Nonparametrics might be a good way to characterise the generalisation abilities of attention-based models, but did not propose a way to do this.}
A DP is a conjugate prior, so we can use exact inference to go from a prior DP to a posterior DP given a set of observations.  The Transformer encoder is used to output a set of pseudo-observations, which parameterise our posterior DP via this inference.  We develop efficient methods to take bounded\footnote{Both sampling and the KL divergence require first assuming an input-dependent bound on samples from both the prior and posterior DPs.  This predefined bound still allows for a variable number of vectors to have nontrivial weights in the mixture distribution.  This will be discussed in Section~\ref{sec:prior}.} samples from this posterior DP, and a reparameterisation trick to allow effective backpropagation through this sampling.  At training time, the Transformer decoder accesses the sampled mixture distribution using query-denoising attention (Figure~\ref{fig:overview} (b)).  At test time, query denoising is applied to the mean of these samples, which is the base distribution of the posterior DP (Figure~\ref{fig:overview} (c)).
The KL divergence between the prior DP and the posterior DP is used to regularise both the information content of individual vectors and the effective number of vectors, giving us our nonparametric variational information bottleneck (NVIB) for Transformer embeddings.

This NVIB regulariser allows us to propose a nonparametric variational autoencoder (NVAE) for Transformer encoder-decoder models, depicted in Figure~\ref{fig:overview} (a).  To evaluate the effectiveness of NVIB for regularising Transformer embeddings, we train a NVAE on natural language text from Wikitext-103 \citep{Merity2017PointerSM} and evaluate the resulting embeddings. Our initial results show that NVIB can be used to create a viable Transformer VAE that is able to reconstruct and generate natural language, indicating that it is able to map the distribution of texts into the prior distribution over latent representations.  Furthermore, NVIB regularises both the effective number of vectors and the amount of information within the vectors. These initial results show the promise of our variational Bayesian framework.

Although our focus is the latent embedding space between Transformer encoders and Transformer decoders, our methods are applicable to any representation which is accessed with attention.
This paper makes the following contributions to learning such attention-based representations.
\begin{itemize}
\item
  Propose Bayesian nonparametrics as a framework for variational Bayesian approaches to attention-based representations, by formalising attention-based representations as mixtures of impulse distributions and generalising attention to Bayesian query denoising (Section \ref{sec:framework}).
\item
  Propose a nonparametric variational information bottleneck (NVIB) regulariser for attention-based representations, using bounded Dirichlet processes for the prior and posterior, and a specific sampling method and its reparameterisation trick (Section \ref{sec:variational}).
\item
  Propose a nonparametric variational autoencoder (NVAE) which is a variational Bayesian version of a Transformer encoder-decoder (Section \ref{sec:nvae}).
\item
  Show that the NVAE model is a viable VAE which can reconstruct and generate, and is able to dynamically regularise both the number and informativeness of component distributions in the latent space. (Section \ref{sec:evaluation}).
\end{itemize}

In the rest of this paper, we start by proposing our nonparametric Bayesian framework for formalising inference of attention-based latent representations (Section~\ref{sec:framework}).  We then present our variational Bayesian approximation and its nonparametric variational information bottleneck (Section~\ref{sec:variational}).  Our NVIB is then used in our deep variational Bayesian model, the nonparametric variational autoencoder (Section~\ref{sec:nvae}).
We evaluate this NVAE model by training it on naturally occurring text and showing that it has the desired characteristics of a VAE and the desired regularisation abilities for Transformer embeddings (Section~\ref{sec:evaluation}).

\section{Nonparametric Bayesian Framework for Attention-based Representations} 
\label{sec:framework}

This section proposes a formalisation of attention-based representations as mixture distributions over a vector space, and propose nonparametric Bayesian methods for modelling information about these mixture distributions.  First we show that standard attention functions can be interpreted as implementing Bayesian denoising, where the set of vectors being accessed specify a mixture of impulse distributions, and the query vector specifies a noisy observation.  We adopt mixture distributions over vectors as a generalisation of attention-based representations, and adopt this denoising function as a generalisation of attention.  Then we use Bayesian nonparametrics to propose prior and posterior distributions over these mixture distributions over vectors.  These priors and posteriors form the basis of our nonparametric variational information bottleneck, proposed in Section~\ref{sec:variational}.

In the rest of this paper we will use the following notation.  Matrices and tensors are uppercase and bold, such as $\vf{A} \in \mathbb{R}^{n\times d}$.
Vectors are lowercase bold and are by default row vectors, such as
$\vf{a} \in \mathbb{R}^{1\times d}$.  Scalars are lowercase italic, $a \in \mathbb{R}$.
For probability distributions,
distributions are uppercase letters, such as $F$, and probability density functions are lowercase letters, such as $f(\cdot)$.
Vectors of mixture distributions are uppercase bold, such as $\vf{F}$.

\subsection{Attention-based Representations as Mixture Distributions for Query Denoising} 
\label{sec:denoising}

When attention is used in a deep learning model, the representation being accessed is treated as a set of vectors.  To access this attention-based representation, a query is given to the attention mechanism which returns the resulting vector.  As the basis of our approach to attention-based representations, we generalise the set of vectors being accessed to a probability distribution over vectors, and generalise the attention function to a function over these probability distributions.

The attention mechanism we assume is scaled dot product attention, standardly used in many attention-based models, including Transformers. For simplicity, we consider cross attention, where a single query vector is mapped to a single result vector.  This attention function uses keys $\vf{K} \in \mathbb{R}^{n \times d}$ and values $\vf{V} \in \mathbb{R}^{n \times d}$ which are both projections from the same set of vectors $\vf{Z} \in \mathbb{R}^{n \times p}$ via weight matrices $\vf{W}^K, \vf{W}^V \in \mathbb{R}^{p \times d}$ respectively.   The query $\vf{q} \in \mathbb{R}^{1\times d}$ is a projection from the input $\vf{u^\prime} \in \mathbb{R}^{1 \times p}$ via the weight matrix $\vf{W}^Q \in \mathbb{R}^{p \times d}$. The keys' dimensionality $d$ is used for scaling.  Scaled dot product attention is then defined as:
\begin{eqnarray*}
  Attention(\vf{u^\prime}, \vf{Z} \;;~ \vf{W}^Q,\vf{W}^k,\vf{W}^Q) 
  &=& \softmax\left( \frac{(\vf{u^\prime} \vf{W}^Q) (\vf{Z} \vf{W}^K)^T}{\sqrt{d}} \right) \vf{Z} \vf{W}^V
  \\\nonumber
  &=& \softmax\left(  \frac{\vf{u} \vf{Z}^T}{\sqrt{d}} \right) \vf{Z} \vf{W}^V,
  \text{~~where~} \vf{u} = \vf{u^\prime} \vf{W}^Q (\vf{W}^K)^T
  \\\nonumber
  &=&  Attn(\vf{u},\vf{Z}) \vf{W}^V
\end{eqnarray*}
where $\vf{u} \in \mathbb{R}^{1\times p}$ is the input query $\vf{u^\prime}$  projected into $\vf{Z}$ space.
In the last line we rewrite scaled dot product attention in terms of a core dot product attention function $Attn(\vf{u},\vf{Z})$ where all operations are done in the space of $\vf{Z}$:
\begin{eqnarray}
  \nonumber
  Attn(\vf{u}, \vf{Z})
  &=& \softmax\left( \tfrac{1}{\sqrt{d}} \vf{u} \vf{Z}^T \right) \vf{Z} 
  \\ \label{eq:attn}
  &=& \sum_{i=1}^n ~\frac{
    \exp(\tfrac{1}{\sqrt{d}} \vf{u} \vf{z}^T_i)
  }{ \sum_{i=1}^n  \exp(\tfrac{1}{\sqrt{d}} \vf{u} \vf{z}^T_i) }~ \vf{z}_i
\end{eqnarray}
where $\vf{z}_i$ is the $i^{\text{th}}$ row of $\vf{Z}$.

In models that use cross attention, an encoder computes the set of vectors $\vf{Z}$, which is the attention-based representation of the input.  A decoder then accesses this representation by issuing queries $\vf{u}$ (or equivalently $\vf{u^\prime}$).  We interpret $\vf{Z}$ as specifying a probability distribution over a vector space, and want to define a function over such probability distributions which, when given any vector $\vf{u}$, always returns the vector $Attn(\vf{u},\vf{Z})$.

\begin{figure}[ht]
  \centering
  \includegraphics[width=0.29\linewidth]{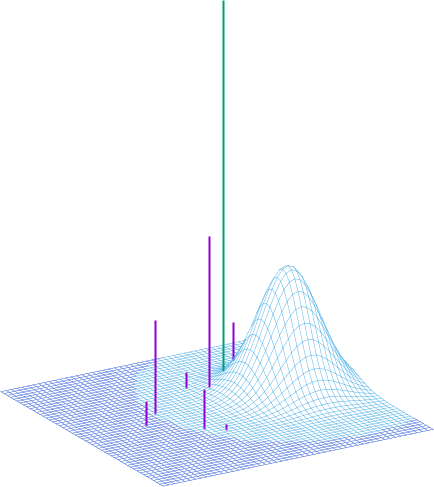}
  \caption{An example of denoising attention.
    Given a \textcolor{purple}{prior distribution} over the true vectors and a \textcolor{lightblue}{noisy observation}, return the \textcolor{green}{expected value} of the denoised vector.}
  \label{fig:denoising}
\end{figure}

We interpret the function $\text{Attn}(\vf{u}, \vf{Z})$ as denoising $\vf{u}$ with a generative model specified by $\vf{Z}$, as depicted in Figure~\ref{fig:denoising}.
The vector $\vf{u}$ is interpreted as an observation of some true vector $\vf{v} \in \mathbb{R}^{1 \times p}$ 
which has been corrupted by Gaussian noise.  The true vector $\vf{v}$ was generated from a prior probability distribution specified by $\vf{Z}$.  The result of $Attn(\vf{u},\vf{Z})$ is the expected value of this true vector $\vf{v}$ after seeing the observation $\vf{u}$, which can be considered the best guess 
of the true vector given the noisy observation, and thus is a form of denoising. 

To derive this interpretation, we interpret the set $\vf{Z}$ of vectors $\vf{z}_i$ as specifying a mixture distribution $F_{\vf{Z}}$ over vectors $\vf{v}$ which consists of one impulse distribution $\delta_{\vf{z}_i}$ at each vector $\vf{z}_i$ weighted by the softmax over their scaled $L_2^2$ norms:
\begin{eqnarray}
  F_{\vf{Z}} &=& \sum_{i=1}^n ~\frac{\exp( \tfrac{1}{2\sqrt{d}}||\vf{z}_i||^2 )}{\sum_{i=1}^n \exp( \tfrac{1}{2\sqrt{d}}||\vf{z}_i||^2 )}~ \delta_{\vf{z}_i}
  \label{eq:probZ}
\end{eqnarray}
Then we can derive this interpretation by replacing attention's sum over $i$ with an integration over $\vf{v}$.
\begin{eqnarray}
  \nonumber
  Attn(\vf{u},\vf{Z})
  &=& \sum_{i=1}^n ~\frac{ \exp(\tfrac{1}{\sqrt{d}} \vf{u} {\vf{z}_i}^T)
  }{ \sum_{i=1}^n  \exp(\tfrac{1}{\sqrt{d}} \vf{u} {\vf{z}_i}^T) }~ \vf{z}_i
  \\ \nonumber
  &=& \sum_{i=1}^n
  ~\frac{ \exp( \tfrac{1}{2\sqrt{d}}||\vf{z}_i||^2 ) ~\exp( {-}\tfrac{1}{2\sqrt{d}}||\vf{z}_i||^2 ) ~\exp(\tfrac{1}{\sqrt{d}} \vf{u} {\vf{z}_i}^T)
  ~ \vf{z}_i
  }{ \sum_{i=1}^n \exp( \tfrac{1}{2\sqrt{d}}||\vf{z}_i||^2 ) ~\exp( {-}\tfrac{1}{2\sqrt{d}}||\vf{z}_i||^2 ) ~\exp(\tfrac{1}{\sqrt{d}} \vf{u} {\vf{z}_i}^T)
  } 
  \\ \nonumber
  &=& \sum_{i=1}^n
  ~\frac{ \exp( \tfrac{1}{2\sqrt{d}}||\vf{z}_i||^2 ) 
  \int_{\vf{v}} \delta_{\vf{z}_i}\hspace{-0.3ex}(\vf{v}) ~\exp( {-}\tfrac{1}{2\sqrt{d}}||\vf{v}||^2 ) ~\exp(\tfrac{1}{\sqrt{d}}\vf{u}\vf{v}^T)
  ~ \vf{v} ~d\vf{v}
  }{ \sum_{i=1}^n \exp( \tfrac{1}{2\sqrt{d}}||\vf{z}_i||^2 )
  \int_{\vf{v}} \delta_{\vf{z}_i}\!(\vf{v}) ~\exp( {-}\tfrac{1}{2\sqrt{d}}||\vf{v}||^2 ) ~\exp(\tfrac{1}{\sqrt{d}}\vf{u}\vf{v}^T) ~d\vf{v}
  }
  \\ \nonumber
  &=& \int_{\vf{v}}
  ~\frac{ \left( \sum_{i=1}^n \frac{\exp( \tfrac{1}{2\sqrt{d}}||\vf{z}_i||^2 )}{\sum_{i=1}^n \exp( \tfrac{1}{2\sqrt{d}}||\vf{z}_i||^2 )}
    \delta_{\vf{z}_i}\!(\vf{v}) \right)
    ~\tfrac{1}{\sqrt{2\pi\sqrt{d}}} \exp({-}\tfrac{1}{2\sqrt{d}} \sum_{k=1}^p ({u}_k - {v}_k)^2)
  }{ \int_{\vf{v}}
    \left( \sum_{i=1}^n \frac{\exp( \tfrac{1}{2\sqrt{d}}||\vf{z}_i||^2 )}{\sum_{i=1}^n \exp( \tfrac{1}{2\sqrt{d}}||\vf{z}_i||^2 )}
    \delta_{\vf{z}_i}\!(\vf{v}) \right)
    ~\tfrac{1}{\sqrt{2\pi\sqrt{d}}} \exp({-}\tfrac{1}{2\sqrt{d}} \sum_{k=1}^p ({u}_k - {v}_k)^2) ~d\vf{v}
  }~ \vf{v} ~d\vf{v}
  \\ \label{eq:genattn}
  &=& \int_{\vf{v}}
  ~\frac{ f_{\vf{Z}}(\vf{v}) ~g(\vf{u}\;;~ \vf{v},\sqrt{d}\vf{I})
  }{ \int_{\vf{v}}  f_{\vf{Z}}(\vf{v}) ~g(\vf{u}\;;~ \vf{v},\sqrt{d}\vf{I}) ~d\vf{v}
  }~ \vf{v} ~d\vf{v}
\end{eqnarray}
where $f_{\vf{Z}}(\cdot)$ is the probability density function for distribution $F_{\vf{Z}}$, and $g(\vf{u}\;;~ \vf{v},\sqrt{d}\vf{I}) = \frac{1}{\sqrt{2\pi\sqrt{d}}} \exp( -\tfrac{1}{2\sqrt{d}} \sum_{k=1}^p ({u}_k-{v}_k)^2 )$ is the multivariate Gaussian function with diagonal variance of $\sqrt{d}$.
The first step just adds terms which don't effect the value.  The second step changes some instances of $\vf{z}_i$ into an integral over $\vf{v}$ which is only nonzero when $\vf{v}{=}\vf{z}_i$ (i.e.\ $\delta_{\vf{z}_i}$).  Thereafter, the terms are rearranged such that the formula reduces to an expected value over $\vf{v}$ with weights proportional to the probability of generating $\vf{v}$ with the distribution $F$ and generating the query $\vf{u}$ with Gaussian noise $\mathcal{N}(\vf{0},\sqrt{d}\vf{I})$ added to $\vf{v}$.\footnote{This derivation is inspired by the interpretation of softmax as Bayesian classification with normally distributed classes \citep{Bishop95}, but here the Gaussian represents uncertainty about the observed vector instead of uncertainty about the class vectors, exploiting the fact that a Gaussian is symmetric in its argument and mean.  This allows us to incorporate the value part of the attention function into a Bayesian denoising interpretation.  To the best of our knowledge, this interpretation of attention is novel.}  Scaling the variance of the multi-dimensional Gaussian noise by $\sqrt{d}$ reduces the impact of the dimensionality $d$ on the similarity $g(\vf{u}\;;~ \vf{v},\sqrt{d}\vf{I})$ between $\vf{u}$ and $\vf{v}$.

In the rest of this paper we will assume the function from equation~\eqref{eq:genattn} as our definition of attention over a probability distribution, which we refer to as \textit{denoising attention}:
\begin{eqnarray}
  \label{eq:dattn}
  DAttn(\vf{u}\;;~ F) &=&
  \int_{\vf{v}} ~\frac{ f(\vf{v}) ~g(\vf{u}\;;~ \vf{v},\sqrt{d}\vf{I})
  }{ \int_{\vf{v}}  f(\vf{v}) ~g(\vf{u}\;;~ \vf{v},\sqrt{d}\vf{I}) ~d\vf{v}
  }~ \vf{v} ~d\vf{v}
\end{eqnarray}
where $f(\cdot)$ is the probability density function for distribution $F$.
The construction above indicates that any scaled dot product attention function is an example of the $ DAttn(\vf{u}\;;~ F)$ function.\footnote{There may be other constructions which could equally well be used to implement $Attn(\vf{u},\vf{Z})$ in terms of $DAttn(\vf{u}\;;~ F)$, but all that is important here is the existence of one.  Equally, we do not intend to claim that all $DAttn(\vf{u}\;;~ F)$ functions can be implemented in terms of $Attn(\vf{u},\vf{Z})$.  Indeed, the greater generality of $DAttn(\vf{u}\;;~ F)$ will be crucial below.
}
However, while attention is only defined over sets-of-vectors $\vf{Z}$, denoising-attention is defined over any probability distribution $F$ over a vector space, not just finite sets of impulse distributions.

\subsection{A Prior over Mixture Distributions from Bayesian Nonparametrics}
\label{sec:prior}

Given that our attention-based latent representations are formalised as distributions $F$, a Bayesian approach requires a prior over these distributions.
Attention-based models place no finite bound on the possible number of vectors in their set of vectors $\vf{Z}$, and thus there is no finite bound on the number of parameters needed to specify the equivalent probability distribution $F$.
Nonetheless, we can still specify distributions $p(F)$ over this infinite space of possible distributions $F$ using methods from Bayesian nonparametrics.  These nonparametric Bayesian methods are specifically designed for modelling probability distributions as unboundedly large mixture models.

\subsubsection{An Infinite Dirichlet Process Prior}
\label{sec:InfDPPrior}

We base our distributions over probability distributions on Dirichlet processes $\DP(G_0,\alpha_0)$.\footnote{Other nonparametric distributions could also be used, but for simplicity we will only consider Dirichlet processes.}  Dirichlet processes (DPs) are a generalisation of Dirichlet distributions to an infinite support, such as the points in a vector space.
A Dirichlet distribution $\Dir(\vf{\alpha})$ is a distribution over probability mass functions $\vf{\pi}$ of discrete categories $i$, $1\leq i\leq \kappa$. We can define $\vf{\pi} \sim \Dir(\vf{\alpha})$ in terms of a probability density function $d(\vf{\pi}\;;~\vf{\alpha})$, defined as:
\begin{eqnarray*}
  d(\vf{\pi}\;;~\vf{\alpha}) &=& \frac{1}{B(\vf{\alpha})} \prod_{i=1}^{\kappa} {\pi}_i^{\alpha_i-1}
\end{eqnarray*}
where here $B(\vf{\alpha})=\frac{\prod_{i=1}^{\kappa} \Gamma({\alpha}_i)}{\Gamma(\sum_{i=1}^{\kappa}{\alpha}_i)}$ is the multivariate beta function, which acts as a normalisation term, and $\Gamma(\cdot)$ is the gamma function.
Samples from a Dirichlet distribution are useful for ``rich-gets-richer'' distributions, where the most probable categories get a large proportion of the probability mass with an exponentially decreasing tail of less probable categories.

There are several equivalent ways to define Dirichlet processes.
One useful definition views a Dirichlet process $F\sim \DP(G_0,\alpha_0)$, where $G_0$ is the base distribution over vectors and $\alpha_0\in \mathbb{R}$ is the concentration parameter, as the limit of a sequence of finite Dirichlet distributions (see \citep{Teh2010}).
\begin{eqnarray*}
  F &=& \sum_{i=1}^\infty {\pi}_i \delta_{\vf{z}_i}
  \\
  \vf{\pi} &\sim& \lim_{\kappa\rightarrow \infty} \Dir(\frac{\alpha_0}{\kappa},\overset{\kappa}{\ldots},\frac{\alpha_0}{\kappa})
  \\
  \vf{z}_i &\sim& G_0 \mbox{~~for~} i=1,\ldots,\infty
\end{eqnarray*}
Note that these are symmetric Dirichlet distributions, in that all the $\kappa$ categories $i$ have the same $\alpha_i=\frac{\alpha_0}{\kappa}$ parameter values.  However, the $\kappa$ categories end up with very different weights ${\pi}_i$, due to the  most probable categories getting a large proportion of the probability mass and the tail of categories getting an exponentially decreasing amount of probability mass. 
In the infinite limit, this tail is infinitely long with infinitesimal probabilities.
The number of categories which get nontrivial probabilities is determined by $\alpha_0$, and becomes independent of $\kappa$ as $\kappa$ gets large.

As shown in this definition,
each sample $F$ from a DP is an infinite mixture of impulse distributions $\delta_{\vf{z}_i}$, parameterised by an infinite sequence of weight-vector pairs ${\pi}_i,\vf{z}_i$.  This distribution $F$ corresponds to the distribution $F_{\vf{Z}}$ in Section~\ref{sec:denoising}, and these weighted vectors ${\pi}_i,\vf{z}_i$ correspond to the elements in the set-of-vectors $\vf{Z}$ with their weights $\softmax( \tfrac{1}{2\sqrt{d}}||\vf{Z}||^2 )$.  The difference is that $F$ is a mixture of an infinite number of weighted vectors, whereas an attention-based representation $\vf{Z}$ is finite.

\subsubsection{The Unbounded Dirichlet Process Prior}
\label{sec:unb-prior}

We do not want a prior which places an apriori bound on the size of $\vf{Z}$, so we assume it is finite but unbounded, and propose a prior which is an unbounded sequence of finite approximations to a DP.  Much work on finite approximations to DPs has used a truncated stick breaking process, but we take an approach which emphasises the exchangeable nature of the distribution generated by a DP.
Our approach is to use the DP definition in Section~\ref{sec:InfDPPrior}, with an unbounded but finite $\kappa$.  We define a distribution over approximations as $\kappa$ increases towards infinity, so that every distribution is over a finite number of vectors, but there is no finite bound on the number of vectors in all distributions.  We define this unbounded DP $F\sim \UDP(G_0,\alpha_0,\phi)$ as an unbounded sequence of finite distributions:
\begin{eqnarray*}
  F &=& \sum_{i=1}^\kappa {\pi}_i \delta_{\vf{z}_i}
  \\
  \kappa &\sim& \phi
  \\
  \vf{\pi} &\sim& \Dir(\frac{\alpha_0}{\kappa},\overset{\kappa}{\ldots},\frac{\alpha_0}{\kappa})
  \\
  \vf{z}_i &\sim& G_0 \mbox{~~for~} i=1,\ldots,\kappa
\end{eqnarray*}
where $\phi$ is some distribution over positive integers $\kappa\in \mathbb{Z}^+$.

Given this distribution over distributions, we also define a conditional distribution over distributions where the number $\kappa$ of vectors is known.  We define this bounded DP $F\sim \BDP(G_0,\alpha_0,\kappa_0)$ as:
\begin{eqnarray*}
  F &=& \sum_{i=1}^{\kappa_0} {\pi}_i \delta_{\vf{z}_i}
  \\
  \vf{\pi} &\sim& \Dir(\frac{\alpha_0}{\kappa_0},\overset{\kappa_0}{\ldots},\frac{\alpha_0}{\kappa_0})
  \\
  \vf{z}_i &\sim& G_0 \mbox{~~for~} i=1,\ldots,\kappa_0
\end{eqnarray*}

We use these definitions both to define a general prior over probability distributions, and to define a conditional prior for each training example.  The general prior is $\UDP(G^p_0,\alpha^p_0,\phi^p)$ where the size distribution $\phi^p$ is determined empirically, as discussed below.\footnote{We will use ``p'' and ``q'' superscripts to designate variables for the prior and posterior, respectively.  Similarly, a zero subscript is part of the name of the variable, in contrast to positive integer subscripts which are indexes.}  The base distribution $G^p_0$ is assumed to be a unit Gaussian, as in previous work \citep{kingma2014autoencoding}.
The concentration parameter $\alpha^p_0$ can be tuned as a hyperparameter, but we set it to one.
\begin{eqnarray*}
  G^p_0 &=& \mathcal{N}(\vf{\mu}^p,D((\vf{\sigma}^p)^2))
  \\
  \alpha^p_0 &=& 1
  \\
  \vf{\mu}^p &=& \vf{0}  
  \\
  \vf{\sigma}^p &=& \vf{1} 
\end{eqnarray*}
where $D(\vf{\sigma})$ maps a vector $\vf{\sigma}$ into a diagonal matrix.

The size of a distribution is viewed as only determining the level of approximation. Hence, we do not try to learn it.  Instead we assume a fixed function of the input text which tells us how many vectors $\kappa_0$ should be included in the approximation.
We then assume the conditional prior $\BDP(G^p_0,\alpha^p_0,\kappa_0)$, which is just the general prior $\UDP(G^p_0,\alpha^p_0,\phi^p)$ where the number of vectors $\kappa_0$ is given.
More specifically, the number $\kappa_0$ of sampled vectors is set to a fixed function of the length of the input text, but does not depend on the actual content of the text.  So the conditional prior $\BDP(G^p_0,\alpha^p_0,\kappa_0)$ is conditioned on the length of the input, but the representation of the content of the text is learned.

\subsubsection{A Conditional Bounded DP Prior}

We generalise this conditioning for the level of approximation to any conditional prior which is a fixed function of only the input length.  Such a conditional prior $\BDP(G^{p^\prime}_0,\alpha^{p^\prime}_0,\kappa_0)$ will be useful below for a softer form of regularisation. 

If we know the input length $n$, but know nothing about the content of the text, then the distribution of vectors should stay the same as the general prior, $G^{p^\prime}_0=G^p_0$.  The number of observations we expect to have after an input of that length would not be $\alpha^p_0$.  Every token can be expected to provide a pseudo-observation count $\Delta^{p^\prime}$ greater than zero, which when added to the unknown pseudo-observation count $\alpha^p_0$ would give us $\alpha^{p^\prime}_0$.  
\[
\alpha^{p^\prime}_0 = \alpha^p_0 + n\Delta^{p^\prime}
\]
This gives us the conditional prior $\BDP(G^p_0,\alpha^{p^\prime}_0,\kappa_0)$.

This conditional prior can be used for regularisation, since it represents knowing nothing about the input other than its length.  It is not used to infer the posterior, discussed below, since in that case we start knowing nothing about the input and only find out about the length after seeing the text.  When computing the KL divergence between the prior and the posterior, we can use the conditional prior instead of the prior.  This means assuming that we do not want to regularise towards a model which gets no information from the input (i.e.\ the prior), but instead regularise towards a model which gets, on average, a minimal amount of information from the input (i.e.\ the conditional prior), similarly to the \emph{free-bits} objective proposed for vector-space VAEs~\citep{kingma2016improved}.  In preliminary experiments, we found that this helps with the stability of training and avoids posterior collapse.

\subsection{A Posterior over Mixture Distributions from Bayesian Nonparametrics}
\label{sec:posterior}

A variational Bayesian model specifies information by specifying a posterior distribution over its representations.  In this section we propose posterior distributions over our attention-based latent representations, which are mixture distributions.  For NVAE, the encoder embeds the input $x$ into the space of posterior distributions $p(G\gvn x)$ over the latent probability distributions $G$.  

To specify this posterior in the output of the encoder, we take advantage of the fact that Dirichlet processes are conjugate priors, meaning that after seeing evidence the posterior distribution is also a DP, and there is a simple formula for inferring this posterior from the prior and a set of observations.  We propose a form for this posterior which allows it to be used in a VAE.  In particular, this posterior allows error backpropagation through the sampling step, and simplifies the computation of the $\KL$ divergence with our prior,
as will be discussed in Section~\ref{sec:variational} on the variational model.

\subsubsection{An Infinite Dirichlet Process Posterior}

In Bayesian terms, the posterior distribution $p(G\gvn x)$ should represent an update to the prior distribution $p(G)$ after seeing some evidence derived from the input $x$.  When we have direct observations $\vf{X}$ of a set of datapoints $\vf{z}_i$, the conjugacy of a DP prior means that the posterior is also a Dirichlet process, and there is a closed-form solution to inferring $p(G\gvn \vf{X})$ from $\vf{X}$ given $p(G)$, namely:
\[
\DP\left( \left(\frac{\alpha^p_0}{\alpha^p_0+|\vf{X}|}G^p_0 + \sum_{i=1}^{|\vf{X}|}\frac{1}{\alpha^p_0+|\vf{X}|}\delta_{\vf{z}_i}\right),~ \left(\alpha^p_0+|\vf{X}|\right) \right).
\]
We can rewrite this in terms of the vocabulary $V(\vf{X})$ of distinct observed vectors and their frequencies $f_\vf{X}(\vf{z}_i)$.
\[
\DP\left( \left(\frac{\alpha^p_0}{\alpha^p_0 +\sum_{i=1}^{|V(\vf{X})|} f_\vf{X}(\vf{z}_i)}G^p_0 + \sum_{i=1}^{|V(\vf{X})|}\frac{f_\vf{X}(\vf{z}_i)}{\alpha^p_0 +\sum_{i=1}^{|V(\vf{X})|} f_\vf{X}(\vf{z}_i)}\delta_{\vf{z}_i}\right),~ \left(\alpha^p_0 +\sum_{i=1}^{|V(\vf{X})|} f_\vf{X}(\vf{z}_i)\right) \right).
\]

We use this conjugacy property to allow the exact inference of a posterior $\DP(G^q_0,\alpha^q_0)$ from a prior $\DP(G^p_0,\alpha^p_0)$ and a set of pseudo-observations estimated by the encoder.  These pseudo-observations are uncertain distributions over observations, which we model with real-valued pseudo-counts ${\alpha}^q_i$ instead of integer frequencies $f_\vf{X}(\vf{z}_i)$, and Gaussian distributions $\mathcal{N}(\vf{\mu}^q_i,D((\vf{\sigma}^q_i)^2))$  instead of exact observations $\delta_{\vf{z}_i}$.  For simplicity, we define the covariance matrix as a diagonal matrix, so that $\vf{\sigma}^q_i$ can be specified as a vector.  A set of $c$ of these parameter triples $\langle{\alpha}^q_i,\vf{\mu}^q_i,\vf{\sigma}^q_i\rangle$ are output by the encoder, 
where ${\alpha}^q_i\in \mathbb{R}_{\geq 0}$, $\vf{\mu}^q_i\in \mathbb{R}^{1 \times d}$ and $\vf{\sigma}^q_i\in \mathbb{R}_{> 0}^{1 \times d}$.
For $1\leq i\leq c$, we define
\begin{eqnarray*}
  {G}^q_i &=& \mathcal{N}(\vf{\mu}^q_i,D((\vf{\sigma}^q_i)^2))
\end{eqnarray*}
as the first $c$ mixture components of the DP's base distribution, with their respective pseudo-counts ${\alpha}^q_i$.

Following the formula for direct observations, we add a $c+1^{\text{th}}$ component to the mixture to represent the ``unknown'' category which the input $x$ does not tell us anything about.  This component has its parameters taken from the prior, namely $\langle\alpha^p_0,\vf{\mu}^p,\vf{\sigma}^p\rangle$.  For notational convenience we define
\begin{eqnarray*}
  {\alpha}^q_{c+1} &=& \alpha^p_0
  \\
  {G}^q_{c+1} &=& G^p_0
  ~~~=~~ \mathcal{N}(\vf{\mu}^p,D((\vf{\sigma}^p)^2))
  ~~=~~ \mathcal{N}(\vf{0},\vf{I})
\end{eqnarray*}

The formula for direct observations can now be applied to the $c$ uncertain observations output by the encoder and the prior for the unknown case, giving us a closed form solution to exact inference of the posterior as the following Dirichlet process:
\begin{eqnarray}
  \label{eq:inf-posterior}
  F &\sim& \DP\left( G^q_0,~ \alpha^q_0 \right)
  \\\nonumber
  G^q_0
  &=& \sum_{i=1}^{c+1} \frac{{\alpha}^q_i}{\alpha^q_0} {G}^q_i 
  ~~~=~~  \frac{\alpha^p_0}{\alpha^q_0} G^p_0 + \sum_{i=1}^{c} \frac{{\alpha}^q_i}{\alpha^q_0} {G}^q_i 
  \\\nonumber
  \alpha^q_0 &=& \sum_{i=1}^{c+1} {\alpha}^q_i
  ~~~=~~ \alpha^p_0 + \sum_{i=1}^{c} {\alpha}^q_i
\end{eqnarray}
We can think of each encoder output ${\alpha}^q_i,\vf{\mu}^q_i,\vf{\sigma}^q_i$ as specifying a cluster of vectors, and the prior as specifying a cluster for unknown vectors.  These clusters are mixed proportionately to their ${\alpha}^q_i$ to get the base distribution.  Samples from this continuous mixture distribution are then taken to get the set of impulses for one of the discrete mixture distributions sampled from the posterior distribution, as depicted above in Figure~\ref{fig:overview}.  In this discrete mixture, most of the weight is concentrated on only a few vectors, as determined by the concentration parameter $\alpha^q_0$.

\subsubsection{A Factorised Dirichlet Process Posterior}
\label{sec:fdp}

As discussed below in Section~\ref{sec:sampling}, naively sampling from the base distribution $G^q_0$ of this posterior DP causes problems for our variational Bayesian model, so we reformulate this DP into an equivalent form which better matches our sampling method.  This form factorises the posterior DP into first generating a total probability mass for each component ${G}^q_i$ of the base distribution, and then generating a mixture distribution from each component ${G}^q_i$.  We call this a \textit{factorised DP}.

Given the vector of components $\vf{G}^q{=}({G}^q_1,\ldots,{G}^q_{c+1})$ of the base distribution $G^q_0$, and their respective pseudo-counts $\vf{\alpha}^q{=}({\alpha}^q_1,\ldots,{\alpha}^q_{c+1})$ which sum to $\alpha^q_0$, we define a factorised DP $F\sim \FDP(\vf{G}^q,\vf{\alpha}^q)$ as:
\begin{eqnarray}
  \label{eq:fac-posterior}
  F &=& \sum_{i=1}^{c+1}~ {\rho}_i {F}_i
  \\ \nonumber
  \vf{\rho} &\sim& \Dir({\alpha}^q_1,\ldots,{\alpha}^q_{c+1})
  \\ \nonumber
  {F}_i &\sim& \DP({G}^q_i,{\alpha}^q_i)
  \mbox{~~for~} i=1,\ldots,c+1
\end{eqnarray}
Each ${\rho}_i$ is the total probability mass
assigned to component $i$ of the base distribution.  Each ${F}_i$ is the normalised mixture of impulse distributions
generated by component $i$.
This reformulation of the posterior is exactly equivalent to the DP posterior in definition~\eqref{eq:inf-posterior}.
\begin{eqnarray*}
  \FDP(\vf{G}^q,\vf{\alpha}^q)
  &=& \DP\left(\left(\sum_{i=1}^{c+1} \frac{{\alpha}^q_i}{\sum_{i=1}^{c+1} {\alpha}^q_i} {G}^q_i\right),~
  \left(\sum_{i=1}^{c+1} {\alpha}^q_i\right)\right)
  ~~=~~ \DP(G^q_0,\alpha^q_0)
\end{eqnarray*}
The proof is given in Appendix~\ref{sec:fdpderivaion}.

This equivalence highlights an important property of Dirichlet processes.  Even though the base distribution has a discrete set of components, so generating from the base distribution seems to involve a discrete choice between components,
an infinite DP does not really model any such discrete choice.
All the components simply generate a countably infinite number of vectors.  The only difference between components is the distribution of weights assigned to their vectors.  As shown in definition~\eqref{eq:fac-posterior}, this weight distribution is determined by the concentration parameter ${\alpha}^q_i$ for that component, plus the Dirichlet distribution across components.  
As discussed in Section~\ref{sec:sampling}, this reformulation from a discrete choice between components $i$ to a continuous choice between weights $\vf{\rho}$ is crucial for our proposed sampling method.

\subsubsection{The Bounded Dirichlet Process Posterior}
\label{sec:bfdp}
  
As with the prior, we approximate the exact posterior over infinite mixtures of impulse distributions with a bounded posterior.  This will be necessary below so that both the samples from the posterior and its KL divergence with the prior can be finite.  In particular, we assume the same number $\kappa_0$ of impulses as for the conditional prior for the given input.  We also assume some method for partitioning these $\kappa_0$ impulses into ${\kappa}_i$ impulses for each of the $c+1$ components of the base distribution.

Given this vector of counts $\vf{\kappa}{=}({\kappa}_1,\ldots,{\kappa}_{c+1})$ such that $\sum_{i=1}^{c+1}{\kappa}_i = \kappa_0$, and the vector of components $\vf{G}^q$ and pseudocounts $\vf{\alpha}^q$ introduced above, we define a bounded version of the above factorised Dirichlet process in an analogous way to the bounded Dirichlet process defined in Section~\ref{sec:unb-prior}.  This gives us our posterior $F\sim \BFDP(\vf{G}^q,\vf{\alpha}^q,\vf{\kappa})$, defined as:
\begin{eqnarray*}
  F &=& \sum_{i=1}^{c+1}~ {\rho}_i {F}_i
  \\
  \vf{\rho} &\sim& \Dir({\alpha}^q_1,\ldots,{\alpha}^q_{c+1})
  \\
  {F}_i &\sim& \BDP({G}^q_i,{\alpha}^q_i,{\kappa}_i)
  \mbox{~~for~} i=1,\ldots,c+1
\end{eqnarray*}
where $\BDP()$ was defined in Section~\ref{sec:unb-prior}.
Expanding out the definition of $\BDP()$, we can also define $F\sim \BFDP(\vf{G}^q,\vf{\alpha}^q,\vf{\kappa})$ as:
\begin{eqnarray}
  \label{eq:eq:bnd-posterior}
  F &=& \sum_{i=1}^{c+1} \sum_{j=1}^{{\kappa}_i}~ {\rho}_i {\pi}^\prime_{ij} \delta_{\vf{z}_{ij}}
  \\ \nonumber
  \vf{\rho} &\sim& \Dir({\alpha}^q_1,\ldots,{\alpha}^q_{c+1})
  \\ \nonumber
  \vf{\pi}^\prime_i &\sim& \Dir(\frac{{\alpha}^q_i}{{\kappa}_i},\overset{{\kappa}_i}{\ldots},\frac{{\alpha}^q_i}{{\kappa}_i} )
  \mbox{~~for~} i=1,\ldots,c+1
  \\ \nonumber
  \vf{z}_{ij} &\sim& {G}^q_i \mbox{~~for~} i=1,\ldots,c+1,~~ j=1,\ldots,{\kappa}_i
\end{eqnarray}
Thus, each probability distribution $F$ sampled from the distribution $\BFDP(\vf{G}^q,\vf{\alpha}^q,\vf{\kappa})$ consists of $\kappa_0$ impulse distributions at vectors $\vf{z}_{ij}$ with weights ${\rho}_i {\pi}^\prime_{ij}$.

\section{The Nonparametric Variational Information Bottleneck}  
\label{sec:variational}

Given the nonparametric prior and posterior from Section~\ref{sec:framework}, we can define our nonparametric variational information bottleneck regulariser by specifying how to compute the KL divergence between prior and posterior, how to effectively sample from the posterior for training, and the mean of these samples to be used at test time.  We start with a review of VIB to provide the context for these contributions, and to motivate the above approach of bounding the DPs.  As far as we are aware, this proposal is the first VIB model for attention-based representations like Transformer embeddings.

\subsection{Variational Information Bottleneck and the ELBO}
\label{sec:elbo}

The idea behind the variational information bottleneck \citep{alemi2017deep} was originally motivated in the context of variational autoencoders by the evidence lower bound (ELBO) \citep{kingma2014autoencoding}.  The ELBO is commonly used in variational Bayesian methods as an objective which approximately maximises the log-likelihood of the observed data $x$.  \citet{kingma2014autoencoding} derive the ELBO as follows.
\begin{eqnarray*}
  \log(p(x)) 
  &=& \E_{q(F\gvn x)} \log(p(x\gvn F)) ~ -\KL(q(F\gvn x)\kld p(F)) ~ +\KL(q(F\gvn x)\kld p(F\gvn x))
  \\
  &\geq& \E_{q(F\gvn x)} \log(p(x\gvn F)) ~ -\KL(q(F\gvn x)\kld p(F)) 
\end{eqnarray*}
The first term of the bound is the reconstruction loss computed from samples $F$ from the estimated posterior $q(F\gvn x)$, and the second term is the KL divergence between the estimated posterior and the prior $p(F)$.
This is a lower bound because $\KL(q(F\gvn x)\kld p(F\gvn x))$ is always positive.  By removing this term, we avoid needing to know the true posterior distribution $p(F\gvn x)$.
This derivation also tells us the size of the error between the bound and the true log-likelihood, namely $\KL(q(F\gvn x)\kld p(F\gvn x))$ (i.e.\ the looseness of the bound).

This analysis helps us understand some of the issues in choosing our prior and posterior, in particular the role of the bounds placed on samples from the prior and posterior.  If both our estimated posterior $q(F\gvn x)$ and the true posterior $p(F\gvn x)$ were infinite Dirichlet processes, then their KL divergence would be infinite.  This explains why we cannot use infinite DPs for our prior and posterior; if we did then the looseness of the ELBO would be infinite.
Our approach is to find a compromise between truncating the prior and posterior DPs at a point where the looseness of the ELBO bound is sufficiently low and where the approximation of the theoretically-motivated infinite posterior is sufficiently accurate (e.g.\ in terms of $L_1$ difference between the generated distributions).
In this work we assume that this point (i.e.\ $\kappa_0$) can be approximated with a fixed linear function of text length, which we treat as a hyperparameter, and leave further investigation to future work.

To motivate its generalisation to models other than autoencoders, the ELBO can be characterised as a variational information bottleneck \citep{alemi2017deep}.
A VIB controls the amount of information which passes from an encoder to a decoder by adding noise to the embedding output by the encoder.  The form of the noise is also output by the encoder, thereby defining a posterior distribution over embeddings (here $q(F\gvn x)$).  A prior distribution defines the uninformative noisy distribution (here $p(F)$), and the KL divergence between the posterior and the prior regularises how much information is conveyed by this posterior distribution (here $\KL(q(F\gvn x)\kld p(F))$).  On the other hand, error backpropagated from the decoder pushes the encoder to include in the posterior all the information needed by the decoder to reconstruct the input (here $p(x\gvn F))$).
The input to the decoder is a sample from the posterior (here $F\,\sim\,q(F\gvn x)$), so we need to backpropagate error from the decoder
through the sampling step.  This is done with a reparameterisation trick \citep{kingma2014autoencoding} whereby all the noise is generated by an unparameterised input.  At test time there is no need to add noise to the embedding, so the mean of the posterior distribution is passed to the decoder (here $\E_{q(F\gvn x)}F$).

This view of the ELBO as a way to regularise the amount of information which passes through a latent representation allows it to be generalised beyond autoencoders.  Under this motivation, there is no need for the different parts of the objective to have the same weight, so the VIB is often generalised to:
\begin{eqnarray*}
  L &=& -\E_{q(F\gvn x)} \log(p(y\gvn F)) ~+~ \lambda~\KL(q(F\gvn x)\kld p(F)) 
\end{eqnarray*}
where the hyperparameter $\lambda$ controls the relative weight of the supervised loss $\E_{q(F\gvn x)} \log(p(y\gvn F))$ and the information bottleneck regulariser $\KL(q(F\gvn x)\kld p(F))$.

In this paper, we split the ELBO into three parts and introduce two hyperparameters to control the relative weight of these parts.
\begin{eqnarray*}
  L &=&  L_R  ~+~ \lambda_D L_D  ~+~ \lambda_G L_G
  \\
  L_R &=& -\E_{q(F\gvn x)} \log(p(x\gvn F))
  \\
  L_D+L_G &=&  \KL(q(F\gvn x)\kld p(F))
\end{eqnarray*}
The two parts for the KL divergence correspond to the KL divergence for the distribution over weights $\vf{\pi}$ generated by the Dirichlet distributions ($L_D$), and the KL divergence for the distribution over vectors $\vf{Z}$ generated by the component Gaussians ($L_G$).  These will be defined in the next subsection.

\subsection{The KL Divergence between the Posterior and Prior}
\label{sec:KL}

Since both the prior and the posterior are conditioned on the same bounds $\vf{\kappa}$ on the number of vectors sampled, we can compute a meaningful finite KL divergence between the prior $\BDP(G^p_0,\alpha^p_0,\kappa_0)$ and the posterior $\BFDP(\vf{G}^q,\vf{\alpha}^q,\vf{\kappa})$, given $\vf{\kappa}$ and $\kappa_0{=}\sum_{i=1}^{c+1}{\kappa}_i$.  We also consider how to approximate the KL divergence in the case where we are only given the overall bound $\kappa_0$, and not the bounds for each component of the posterior $\vf{\kappa}$.

To directly compare the posterior with the prior, we first reformulate the prior as a factorised DP with the same form as the posterior.  We can do this without changing the distribution specified by the prior, simply by making $c+1$ copies of the base distribution $G^p_0$ and weighting those copies proportionately to the weights $\frac{{\alpha}^q_i}{\alpha^q_0}$ of the components of the posterior base distribution.  This gives us the prior $\BFDP(\vf{G}^p,\vf{\alpha}^p,\vf{\kappa})$ where $\vf{G}^p = (G^p_0,\overset{c+1}{\ldots},G^p_0)$ and $\vf{\alpha}^p = \vf{\alpha}^q\frac{\alpha^p_0}{\alpha^q_0} = (\alpha^p_0\frac{{\alpha}^q_1}{\alpha^q_0},\overset{c+1}{\ldots},\alpha^p_0\frac{{\alpha}^q_{c+1}}{\alpha^q_0})$.

The derivation of the KL divergence is given in Appendix~\ref{sec:KLderivation}.  It has two terms, one for the distribution of weights $\vf{\pi}$ generated by the Dirichlet distributions ($L_D$), and one for the distribution of vectors $\vf{Z}$ generated by the component Gaussians ($L_G$), as introduced above.
\begin{eqnarray}
  \label{eq:KLloss}
  &&\hspace{-8.5ex} 
  \KL( \BFDP(\vf{G}^q,\vf{\alpha}^q,\vf{\kappa})
  \kld \BFDP(\vf{G}^p,\vf{\alpha}^q\tfrac{\alpha^p_0}{\alpha^q_0},\vf{\kappa}) )
  ~~=~~ L_D ~+~ L_G
  \\ \nonumber
  L_D
  &\!=\!& \log\Gamma(\alpha^q_0) -\log\Gamma(\alpha^p_0)
  +(\alpha^q_0-\alpha^p_0) \left(
  -\psi(\alpha^q_0) +\sum_{i=1}^{c+1} \tfrac{{\alpha}^q_i}{\alpha^q_0} \psi(\frac{{\alpha}^q_i}{{\kappa}_i})
  \right)
  +\sum_{i=1}^{c+1} {\kappa}_i \left( \log\Gamma(\frac{\alpha^p_0{\alpha}^q_i}{\alpha^q_0{\kappa}_i}) -\log\Gamma(\frac{{\alpha}^q_i}{{\kappa}_i}) \right)
  \\ \nonumber
  L_G
  &\!=\!& 
  \tfrac{1}{2} \sum_{i=1}^{c+1} {\kappa}_i \sum_{h=1}^d \left(
  \frac{({\mu}^q_{ih}-{\mu}^p_{h})^2}{({\sigma}^p_{h})^2} + \frac{({\sigma}^q_{ih})^2}{({\sigma}^p_{h})^2}
  -1 -\log\frac{({\sigma}^q_{ih})^2}{({\sigma}^p_{h})^2} \right)
\end{eqnarray}
where $\psi$ is the digamma function and again $\Gamma$ is the gamma function.

Equation \eqref{eq:KLloss} assumes we are given the bounds ${\kappa}_i$ for each component $i$ of the posterior, but what if we are only given the bound $\kappa_0$ for the total number of vectors sampled.  In this case we would ideally like to marginalise over $\vf{\kappa}$
given $\kappa_0$, but this doesn't appear to be tractable for $L_D$.  In this case, we approximate the expectation of $L_D$ over $\vf{\kappa}$ with the $L_D$ of the expectation of $\vf{\kappa}$, meaning that we simply substitute the expected value of $\vf{\kappa}$ for $\vf{\kappa}$ in equation \eqref{eq:KLloss}.
This approximation is justified by the fact that both $\psi(\tfrac{{\alpha}^q_i}{{\kappa}_i})$ and
${\kappa}_i \left( \log\Gamma(\frac{{\alpha}^q_i}{\alpha^q_0 {\kappa}_i}) -\log\Gamma(\frac{{\alpha}^q_i}{{\kappa}_i})  \right)$ are approximately linear in ${\kappa}_i$ for large ${\kappa}_i$ (see Appendix~\ref{sec:KLderivation}).  
\begin{eqnarray}
  \label{eq:KLloss-kappa}
  {\kappa}_i &=& \frac{{\alpha}^q_i}{\alpha^q_0} \kappa_0
  \\ \nonumber
  L_D
  &\approx& \log\Gamma(\alpha^q_0) -\log\Gamma(\alpha^p_0)
  +(\alpha^q_0-\alpha^p_0) \left(
  -\psi(\alpha^q_0) +\psi(\frac{\alpha^q_0}{\kappa_0})
  \right)
  +\kappa_0 \left( \log\Gamma(\frac{\alpha^p_0}{\kappa_0}) -\log\Gamma(\frac{\alpha^q_0}{\kappa_0}) \right)
  \\ \nonumber
  L_G
  &=& 
  \tfrac{1}{2} \kappa_0 \sum_{i=1}^{c+1} ~\frac{{\alpha}^q_i}{\alpha^q_0}~  \sum_{h=1}^d \left(
  \frac{({\mu}^q_{ih}-{\mu}^p_{h})^2}{({\sigma}^p_{h})^2} + \frac{({\sigma}^q_{ih})^2}{({\sigma}^p_{h})^2}
  -1 -\log\frac{({\sigma}^q_{ih})^2}{({\sigma}^p_{h})^2} \right)
\end{eqnarray}
As with ${\kappa}_i$, both $\psi(\frac{\alpha^q_0}{\kappa_0})$ and
$\kappa_0 \left( \log\Gamma(\frac{\alpha^p_0}{\kappa_0}) -\log\Gamma(\frac{\alpha^q_0}{\kappa_0}) \right)$ are approximately linear in $\kappa_0$ for large $\kappa_0$.  Thus both $L_D$ and $L_G$ scale approximately linearly with $\kappa_0$.

\subsection{Sampling a Mixture Distribution from the Posterior}
\label{sec:sampling}

To enforce a probabilistic interpretation of the posterior parameters output by the encoder and control the amount of information which passes to the decoder,
at training time a VAE \citep{kingma2014autoencoding} samples from this posterior distribution and uses this sample to reconstruct the input.  The ``reparameterisation trick'' is used to ensure that backpropagation of the reconstruction error through this sampling step can be done effectively.  The idea of the reparameterisation trick is to push the randomness used for sampling into an unparameterised input variable.

We propose a novel reparameterisation trick for bounded Dirichlet processes which allows sampling without any categorical choices, and propose specific sampling methods which result in effective backpropagation through the sampling step.

\subsubsection{A Factorised Sampling Method}

For our NVIB model, we sample the parameters $\langle\vf{\pi},\vf{Z}\rangle$ of a mixture distribution $F$ generated by our bounded Dirichlet process posterior $\BFDP(\vf{G}^q,\vf{\alpha}^q,\vf{\kappa})$, where $F$ consists of a set of impulse distributions $\delta_{\vf{z}_k}$ each with a weight ${\pi}_k$.  
A straightforward approach to sampling from a Dirichlet process would sample weights $\vf{\pi}$ from a Dirichlet distribution and sample vectors $\vf{Z}$ from the base distribution of the DP, where sampling from the base distribution involves first sampling a component of the base distribution and then sampling a vector from that component's Gaussian.  However, this sampling method has the problem that there is no reparameterisation trick for the categorical choice between the components of the base distribution.

To solve this problem, we take advantage of the insight from Section~\ref{sec:fdp} that a DP does not require any such categorical choice.  Rather than doing a sequence of discrete choices between components of the base distribution, the factorisation of a DP given in Section~\ref{sec:fdp} allows us to do a continuous choice of a set of weights, each one corresponding to the total weight for one component.  These total weights $\vf{\rho}$ are then combined with individual weights $\vf{\pi}^\prime$ and vectors $\vf{Z}$ sampled from each component separately.  

Since our components use Gaussian distributions over vectors, we can use the same reparameterisation trick as \citet{kingma2014autoencoding} for sampling vectors $\vf{Z}$ from an individual component.
We sample the weights $\vf{\pi}$ directly
as specified in the definition of our $\BFDP$ posterior in Section~\ref{sec:bfdp}, namely by sampling the total weights $\vf{\rho}$ across components and sampling the weights $\vf{\pi}^\prime_i$ for individual vectors within a component $i$.  Both these sampling steps sample a normalised set of weights from a Dirichlet distribution.  The fact that our posterior is over a bounded number $\kappa_0$ of vectors ensures that all of these Dirichlet distributions are over a finite set.
For the total weights, the Dirichlet distribution is over weights across the set of components, which is a bounded set for any given posterior.  For the individual components, the Dirichlet distribution is over weights across the set of ${\kappa}_i$ vectors to be sampled from that component, given in the set of bounds $\vf{\kappa}$ for the posterior $\BFDP(\vf{G}^q,\vf{\alpha}^q,\vf{\kappa})$.

This leaves the question of how to sample from a Dirichlet distribution and what reparameterisation trick to use for backpropagation through this sample.  This cannot be done exactly, but there are good approximations, as discussed below.

\subsubsection{Sampling Vectors from a Component of the Base Distribution}

Each vector $\vf{z}_k$ is sampled independently from some specific component ${G}^q_i$ of the base distribution $G^q_0$.  Since we assume that all these components are distributed according to a Gaussian
${G}^q_i = \mathcal{N}(\vf{\mu}^q_i,D((\vf{\sigma}^q_i)^2))$,
we can sample from this distribution using location-scale shifting \citep{kingma2014autoencoding}:
\begin{eqnarray}
  \vf{z}_k &=& \vf{\mu}^q_i + \vf{\sigma}^q_i \epsilon_k
  \\ \nonumber
  \epsilon_k &\sim& \mathcal{N}(\vf{0},\vf{1})
\end{eqnarray}
Since the random sampling comes from an unparameterised unit Gaussian, there is no need to backpropagate error into this sampling step, but the error can be backpropagated into $\vf{\mu}^q_i$ and $\vf{\sigma}^q_i$ given a specific sample.  This is the reparameterisation trick for Gaussian distributions.

\subsubsection{Sampling from a Dirichlet Distribution}

A Dirichlet distribution over category weights can be sampled by sampling from a Gamma distribution for each category and then normalising.  
A sum-normalised set of $\kappa$ random variables $\pi_1,\ldots,\pi_\kappa \sim Dir(\alpha_1,\ldots,\alpha_\kappa)$ follows a Dirichlet distribution if the unnormalised random variables $\gamma_i$ each follow a Gamma distribution. 
\begin{eqnarray}
  \pi_i &=& \frac{\gamma_i}{\sum_i^\kappa \gamma_i}
  \\ \nonumber
  \gamma_i &\sim& \Gamma(\alpha_i,\beta{=}1)
\end{eqnarray}
where $\Gamma(\alpha_i,\beta{=}1)$ is the Gamma distribution with $\beta{=}1$, whose PDF is
$f(x) = \frac{1}{\Gamma(\alpha_i)}\exp((\alpha_i-1)\log(x)-x)$.
There is no reparameterisation trick for the exact Gamma distribution, but there are for approximations.
We propose to use a combination of two approximations for the Gamma distribution which have a reparameterisation trick, one for small values of $\alpha_i$ and one for larger values of $\alpha_i$.

\paragraph{Inverse CDF approximation to Gamma distributions.}
The Gamma distribution cannot use location-scale shifting for sampling due to its asymmetry, nor can the parameters and noise components be decoupled in the inverse CDF. Hence, \citep{knowles2015stochastic} suggests sampling using an approximation to the inverse CDF of the Gamma distribution of the following form:
\begin{eqnarray}
  \gamma_i &\approx& \beta^{-1} (u_i \alpha_i \Gamma(\alpha_i))^{1 / \alpha_i},
  \\ \nonumber
  u_i &\sim& \text{U}(0,1).
\end{eqnarray}
This approximation allows the inverse CDF of the Gamma distribution to be a function of the parameters and independent noise from a uniform distribution $\text{U}(0,1)$. However, this approximation is only recommended when the value of $\alpha_i < 1$ and $\beta =1$.  In our case $\beta =1$ but we sometimes have large $\alpha_i$.

\paragraph{Gaussian approximation to Gamma distributions.}
\citep{knowles2015stochastic} further mentions that the Gaussian distribution can be used to approximate a Gamma distribution for larger $\alpha$.
The Gaussian distribution is a symmetric distribution which can be sampled by location-scale shifting, as discussed above. 
The Gamma distribution $\Gamma(\alpha,\beta{=}1)$ can be approximated with a Gaussian of the form $\gamma \sim \mathcal{N}(\alpha,\sqrt{\alpha})$ \citep{knowles2015stochastic}, which gives us:
\begin{eqnarray}
  \gamma_i &\approx& \alpha_i + \sqrt{\alpha_i} \epsilon_i
  \\ \nonumber
  \epsilon_i &\sim& \mathcal{N}(0,1)
\end{eqnarray}
The Gaussian distribution is symmetric and can take on negative values. Hence, this approximation is inappropriate for the Gamma distribution unless the $\alpha_i$ parameter is sufficiently large, otherwise the sample will need to be truncated to a value greater than zero.

\begin{figure}[ht]
  \centering
  \includegraphics[width=0.54
    \textwidth]{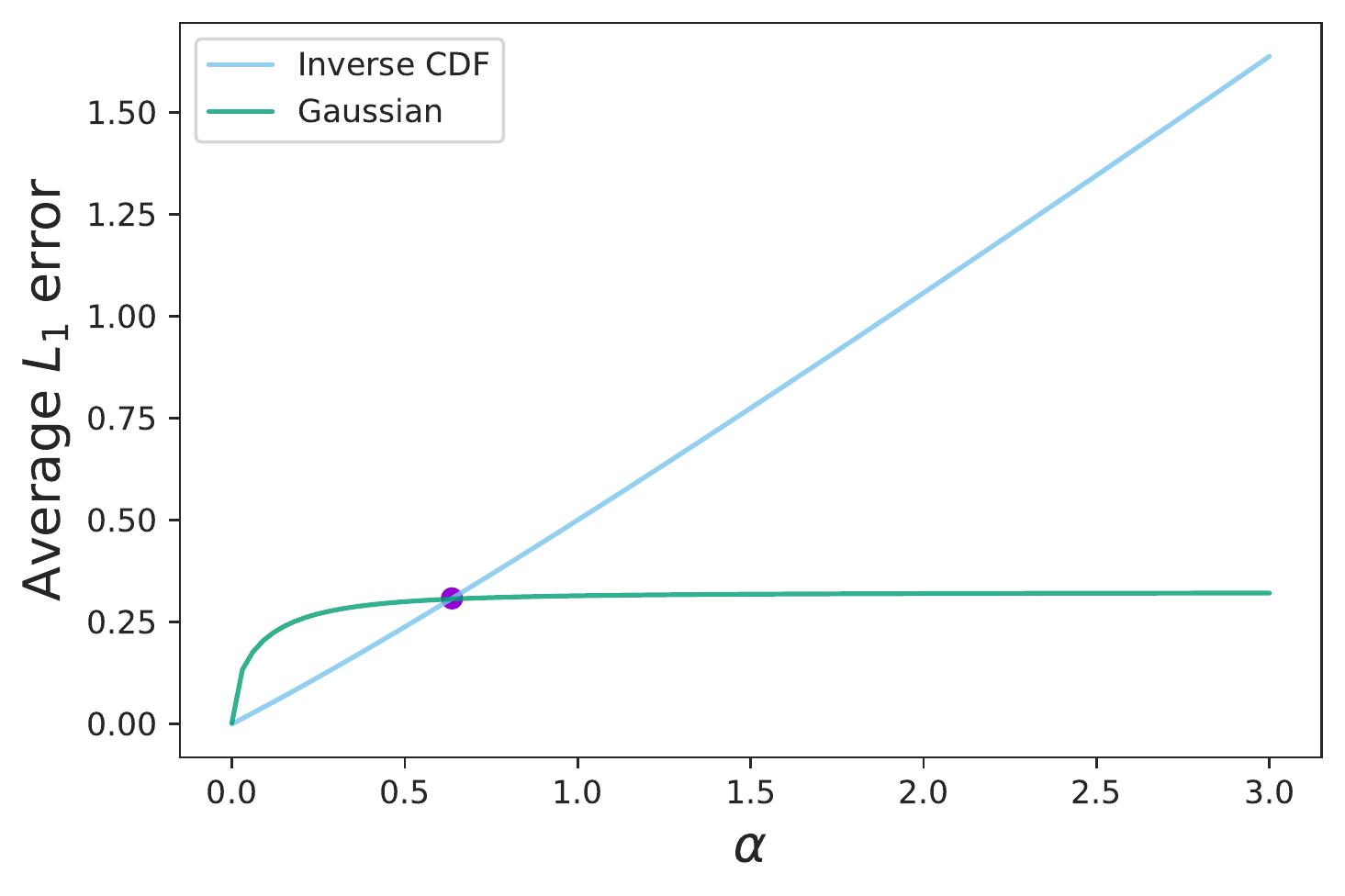}
  \caption{The average absolute difference between a Gamma inverse CDF function and our approximations: \textcolor{lightblue}{inverse CDF approximation} and \textcolor{green}{Gaussian inverse CDF} are plotted for values of $\alpha$.}
  \label{fig:approximation}
\end{figure}

\paragraph{The combined reparameterisation of Gamma distributions.}
To visualise the error for these two approximations, their average $L_1$ distance from the true Gamma inverse CDF is plotted in Figure~\ref{fig:approximation}
for different values of $\alpha$. The plot shows that the approximation error is equal when $\alpha = 0.6363$.  To take advantage of the strengths of both these approximations, we propose to reparameterise the Gamma distribution as a blend of these two approximations with a switch at $\alpha = 0.6363$.

\subsection{The Mean Probability Distribution at Test Time}
\label{sec:testtime}

\begin{figure}[t]
  \centering
  \includegraphics[width=0.29\linewidth]{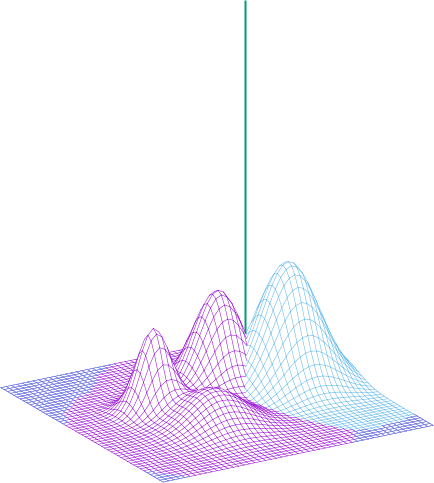}
  ~\raisebox{4ex}{vs}~\includegraphics[width=0.29\linewidth]{figures/denoising_1.png}
    \caption{An example of denoising attention at test time, on left.  Given the \textcolor{purple}{mean distribution} over true vectors and a \textcolor{lightblue}{noisy observation}, return the \textcolor{green}{expected value} of the denoised vector.
      The resulting vector is similar to the \textcolor{green}{vector} computed from a sampled distribution, on right, repeated from Figure~\ref{fig:denoising}.}
    \label{fig:testtime}
\end{figure}

In a fixed-length-vector VAE, instead of sampling at test time, the mean vector is used.  To the trained model, the mean vector looks like a typical vector used at training time.  The mean of a DP is its base distribution, which is also true of our $\BFDP$ posterior.  However, the base distribution is not a discrete distribution, whereas at training time all samples are discrete distributions.  This is one of the main advantages of generalising the attention function to denoising attention, proposed in Section~\ref{sec:denoising}.  Denoising attention can equally well be applied to the mixture of Gaussians of the base distribution as to the discrete distribution from sampling, as depicted in Figure~\ref{fig:testtime}.  On the other hand, the standard attention function is only defined for a finite set of vectors, so even if we convert the sampled discrete distributions to finite sets of vectors, at testing time there is no finite set of vectors which is the mean of the sampled sets of vectors.  In particular, the set of mean vectors is not the mean of the set of vectors, since it underestimates the variance.

Despite the fact that a sample is a discrete distribution and this mean is a continuous distribution, to denoising attention this mixture of Gaussians looks like a typical sample from the posterior, as desired.  We can see this in Figure~\ref{fig:testtime} by comparing the vector returned by denoising attention (in green) given the continuous mean distribution (left) and a typical sample from this distribution (right).  Thus, the function defined by applying denoising attention to a sample distribution can be seen as a noisy version of the function defined by applying denoising attention to the mean distribution.

This is the essence of what NVIB is doing.  Our latent representation is a parameterisation of an attention function from query vectors to result vectors.  The encoder outputs two things, a mean function and a noise level, parameterised by $G^q_0$ and $\alpha_0$ respectively.  At training time we sample noisy versions of the mean function where the amount of noise is determined by $\alpha_0$.  At test time we just use the mean function determined by $G^q_0$.

To efficiently compute denoising attention applied to a mixture of Gaussians, we take advantage of the fact that a multiplication of Gaussians is a Gaussian.  
\begin{eqnarray}
  DAttn(\vf{u}\;;~ G^q_0)
  \nonumber
  &=& \int_\vf{v} ~\frac{
    \left( \sum_i \frac{{\alpha}^q_i}{\sum_i {\alpha}^q_i} g(\vf{v}\;;~ \vf{\mu}^q_i,D((\vf{\sigma}^q_i)^2)) \right) ~ g(\vf{v};\vf{u},\sqrt{d}\vf{I})
  }{\int_\vf{v} \left( \sum_i \frac{{\alpha}^q_i}{\sum_i {\alpha}^q_i} g(\vf{v}\;;~ \vf{\mu}^q_i,D((\vf{\sigma}^q_i)^2)) \right) ~ g(\vf{v};\vf{u},\sqrt{d}\vf{I}) ~d\vf{v}
  } ~\vf{v} ~d\vf{v}
  \\ \nonumber
  &=& \int_\vf{v} ~\frac{
    \sum_i \frac{{\alpha}^q_i}{\sum_i {\alpha}^q_i} g(\vf{u}\;;~ \vf{\mu}^q_i,D({\sqrt{d}+(\vf{\sigma}^q_i)^2}))
    ~g( \vf{v}; (\frac{\tfrac{1}{\sqrt{d}}\vf{u}+\tfrac{1}{(\vf{\sigma}^q_i)^{2}}\vf{\mu}^q_i}{\tfrac{1}{\sqrt{d}}+\tfrac{1}{(\vf{\sigma}^q_i)^{2}}}), D(\frac{1}{\tfrac{1}{\sqrt{d}}+\tfrac{1}{(\vf{\sigma}^q_i)^{2}}}) )
  }{\int_\vf{v} \sum_i \frac{{\alpha}^q_i}{\sum_i {\alpha}^q_i} g(\vf{u}\;;~ \vf{\mu}^q_i,D({\sqrt{d}+(\vf{\sigma}^q_i)^2}))
    ~g( \vf{v}; (\frac{\tfrac{1}{\sqrt{d}}\vf{u}+\tfrac{1}{(\vf{\sigma}^q_i)^{2}}\vf{\mu}^q_i}{\tfrac{1}{\sqrt{d}}+\tfrac{1}{(\vf{\sigma}^q_i)^{2}}}), D(\frac{1}{\tfrac{1}{\sqrt{d}}+\tfrac{1}{(\vf{\sigma}^q_i)^{2}}}) ) ~d\vf{v}
  } ~\vf{v} ~d\vf{v}
  \\
  \label{eq:TestDAttn}
  &=& \sum_i \frac{ 
    {\alpha}^q_i g(\vf{u}\;;~ \vf{\mu}^q_i,D({\sqrt{d}+(\vf{\sigma}^q_i)^2}))    
  }{\sum_i {\alpha}^q_i g(\vf{u}\;;~ \vf{\mu}^q_i,D({\sqrt{d}+(\vf{\sigma}^q_i)^2}))
  } \left( \frac{\tfrac{1}{\sqrt{d}}\vf{u}+\tfrac{1}{(\vf{\sigma}^q_i)^{2}}\vf{\mu}^q_i}{\tfrac{1}{\sqrt{d}}+\tfrac{1}{(\vf{\sigma}^q_i)^{2}}} \right)
\end{eqnarray}
where $D((\vf{\sigma})^2)$ maps a vector into a diagonal matrix, and all algebraic calculations over vectors of $\sigma$s
are done componentwise (i.e.\ independently on each dimension).
Note the similarity of this function to the standard attention function from equation~\eqref{eq:attn}, only here attention is being applied to a set of multivariate Gaussians instead of a bag of vectors.
This attention function is similar to attention over the component means $\vf{\mu}^q_i$, but the scaling is by $\frac{1}{\sqrt{d}+(\vf{\sigma}^q_{i})^2}$, and the attention values are an interpolation between each mean and the query.

\section{The Nonparametric Variational Autoencoder}
\label{sec:nvae}

\begin{figure}[t]
  \centering
  \includegraphics[width=0.63\linewidth]{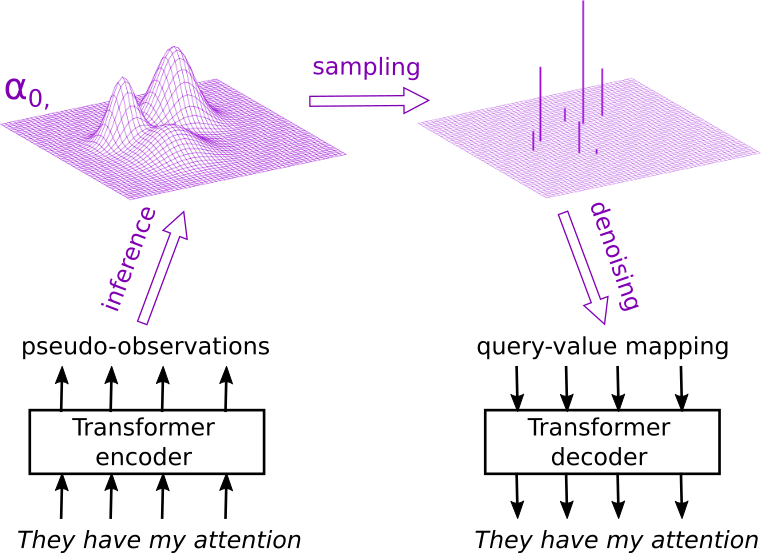}
  \caption{A depiction of the NVAE model, with its \textcolor{purple}{NVIB layer} (in purple).  A Transformer encoder embeds the sentence in the space of Dirichlet Processes, specified by a concentration parameter $\alpha^q_0{=}\sum_i {\alpha}^q_i$, plus a set of Gaussian mixture components parameterised by $\langle{\alpha}^q_i,\vf{\mu}^q_i,\vf{\sigma}^q_i\rangle$.  A discrete sample $F$ adds noise to the mixture distribution according to the concentration parameter.  A Transformer decoder then accesses this sample $F$ using query-denoising attention, and reconstructs the sentence.  NVIB regularises the concentration parameter and the number and informativeness of the components in the mixture distribution.}
  \label{fig:architecture}
\end{figure}

We define a VAE for Transformers by using the nonparametric VIB defined in Section~\ref{sec:variational} to regularise the attention-based representation between the encoder and decoder of a Transformer autoencoder, as depicted in Figure~\ref{fig:architecture}.  
In this NVAE model, the Transformer encoder is used to estimate the parameters $\langle\vf{\alpha}^q,\vf{{\mu}}^q,\vf{{\sigma}}^q \rangle$ of the posterior given the input text $x$.  The Transformer decoder is used to reconstruct the input text $x$ using denoising attention over a sample $F$
from this posterior.  Our NVIB layer regularises the amount of information which passes from the encoder to the decoder through this posterior.

\subsection{The Transformer Encoder}
\label{sec:encoder}

The Transformer encoder maps the input text $x$ to the parameters $\langle\vf{\alpha}^q,\vf{\mu}^q,\vf{\sigma}^q\rangle$ of the Dirichlet process posterior.
Given a text $x$ with $n$ number of tokens,
a standard Transformer encoder is used to compute a vector for each token $i$ of the input.  From each of these $n$ individual token embeddings, the encoder then linearly projects to three parameters, ${\alpha}^q_i\in \mathbb{R}$, $\vf{\mu}^q_i \in \mathbb{R}^{1 \times d} $ and $\log(\vf{\sigma}^q_i) \in \mathbb{R}^{1 \times d}$. The variance parameters are exponentiated to be strictly positive, whereas the pseudo-count parameters ${\alpha}^q_i$ are estimated using a Rectified Linear Unit (ReLU) activation, which results in masking the vector during cross-attention when it is exactly zero.  Thus, the DP posterior has one component $\langle{\alpha}^q_i,\vf{\mu}^q_i,\vf{\sigma}^q_i\rangle$ of its base distribution for each token of the input.

\subsection{The NVIB Regulariser}
\label{sec:bottleneck}

The NVIB layer regularises the amount of information which passes from the encoder to the decoder through the latent mixture distribution $F$.  As discussed in Sections~\ref{sec:elbo} and~\ref{sec:sampling}, at training time the NVIB layer controls the amount of information by adding noise to this base distribution by sampling a discrete distribution $F$ from the DP posterior. This amount of information is regularised with the training loss by including the KL divergence between the posterior and the conditional prior, as discussed in Section~\ref{sec:KL}.  The conditional prior is conditioned on the length $n$ but is otherwise uninformative about the input text $x$.  Finally, in Section~\ref{sec:testtime}, during test time the NVIB layer simply outputs the base distribution $G_0$ of the DP posterior.

As with VIB for vector spaces, the KL divergence encourages the encoder to output component Gaussians with smaller $\vf{\mu}^q_i$ and larger $\vf{\sigma}^q_i$.  With NVIB, the KL divergence also encourages the encoder to output smaller and sparser ${\alpha}^q_i$, which regularises the effective number of components in the posterior as well as the precision of their weights.

In the current version of our NVIB layer we take an efficient and simplified approach to sampling from the posterior $\BFDP(\vf{G}^q,\vf{\alpha}^q,\vf{\kappa})$ by setting ${\kappa}_i{=}1$ for all components $i$.  This means that we only sample a single vector from each Gaussian component.  This simplifies sampling because there is no need to sample a Dirichlet distribution to get the relative weights of different vectors sampled from the same component Gaussian.  We leave the investigation of the case where multiple vectors are sampled per component to future work.  Thus, a single sample consists of $n$ vectors  $\textbf{Z} \in \mathbb{R}^{n \times d}$ and $n$ weights  $\boldsymbol{\pi} \in \mathbb{R}^{n \times 1}$.

This approach to sampling vectors implies that the hyperparameter $\kappa_0$ is set to the sentence length, $\kappa_0{=}n$.  This implies that the $\KL$ divergence losses $L_D$ and $L_G$ grow linearly with the sentence length.  Preliminary experiments suggest that training converges better without this linear dependence, so we set the hyperparameters $\lambda_D$ and $\lambda_G$ to be linear in $\frac{1}{n}$, thereby cancelling out the linear dependence on the sentence length.  In addition, to bring $L_G$ into the same range of values as $L_D$, we scale the hyperparameter $\lambda_G$ by $\frac{1}{d}$, removing the dependence on the dimensionality $d$ of vectors.
\begin{eqnarray*}
\lambda_D &=& \frac{1}{n} \lambda^\prime_D
\\
\lambda_G &=& \frac{1}{d} \frac{1}{n} \lambda^\prime_G
\end{eqnarray*}

\subsection{The Transformer Decoder}
\label{sec:decoder}

The Transformer decoder receives a distribution $F$ over vectors and reconstructs the input text $x$.
During training, $F$ is specified by the sampled vectors $\textbf{Z} \in \mathbb{R}^{n \times d}$ and the sampled weights $\vf{\pi} \in \mathbb{R}^{n \times 1}$, and at test time $F$ is specified by the output of the encoder $\vf{\alpha}^q \in \mathbb{R}^{n \times 1}$, $\vf{\mu}^q \in \mathbb{R}^{n \times d}$ and $\text{log}(\vf{\sigma})^q \in \mathbb{R}^{n \times d}$.  In both cases, the decoder accesses $F$ using denoising attention in the same way that standard Transformer decoders use cross attention.
During training, the text is predicted using teacher forcing, and during test time the text is predicted autoregressively using greedy search until the end of sequence token is generated or the sentence generated is twice the target length.

\subsection{The Resulting Generative Model}
\label{sec:generation}

One of the interesting characteristics of a VAE is that it can be used as a generative model for examples from the training distribution.  A well-trained encoder will project the entire training distribution into the entire prior distribution, and a well-trained decoder will project these points in the prior distribution back to examples in the training distribution.  With appropriate regularisation, the decoder mapping should also be smooth, meaning that any point sampled from the prior distribution will be mapped to examples that look like they come from the training distribution, even if they did not appear in the training data.

To use our NVAE model as a generative model, we sample from the prior and use the trained Transformer decoder to generate a sentence.  As discussed in Section~\ref{sec:prior}, to sample from the same prior as used for training, we need to first sample a sentence length, and then sample from the conditional prior given that sentence length.  For simplicity, we sample the sentence length from the empirical distribution of sentence lengths in the training data.

\section{Intrinsic Evaluations of NVIB in a NVAE}
\label{sec:evaluation}

In this paper we report initial results on the regularisation abilities of NVIB when training a NVAE on natural text.
These experiments are designed to test two hypotheses:
\textbf{H1} the NVAE Transformer model is a viable VAE in that it is able to both reconstruct sentences and generate sentences by sampling from the prior;
\textbf{H2} the NVIB layer is able to dynamically regularise both the number and informativeness of component distributions in the latent space of mixture distributions.
We compare these abilities to a variety of alternative Transformer-based autoencoders.

Showing that the induced representations are useful in specific tasks (extrinsic evaluation) is currently being left for future work.

\subsection{Experimental Setup}

\subsubsection{Metrics}
\label{sec:metrics}

These experiments focus on the reconstruction and generation abilities of the model. For reconstruction, we report SacreBleu \citep{post-2018-call} and perplexity scores as measures of the model's ability to reconstruct sentences from unseen validation data.   For generation, we report two automatic metrics, namely forward perplexity (F-PPL) and reverse perplexity (R-PPL) \citep{DBLP:journals/corr/ZhaoKZRL17, cifka2018eval}, which both measure how well the generated distribution matches the data distribution.

The two generation metrics both involve training an external language model, for which we use a standard auto-regressive Transformer language model.  First we  generate 100k sample sentences from the model we wish to evaluate. 
The forward perplexity F-PPL measure is the perplexity of the external language model trained on training data and evaluated on the generated text. This measures the fluency of the generated text, but cannot detect the collapsed case where the model repeatedly generates a few common sentences \citep{DBLP:journals/corr/abs-1905-12777}. The reverse perplexity R-PPL measure is the perplexity of the external language model trained on generated samples and evaluated on validation data. This reflects fluency but prioritises the diversity of the generated text.  If the model being evaluated generates only a few common sentences, the language model trained on these samples will exhibit good F-PPL, but poor R-PPL on validation data \citep{DBLP:journals/corr/ZhaoKZRL17}. If the model being evaluated generates non-fluent sentences with high word diversity then these samples will exhibit better R-PPL, but poor F-PPL.

\subsubsection{Data}
The dataset Wikitext-103 \citep{Merity2017PointerSM} was selected as it is a general English language corpus containing good and featured Wikipedia articles. It contains roughly $4$ million training sentences with the longest sentences having more than $100$ subword tokens. The BERT-base, uncased tokeniser is used for tokenisation as it is a generally accessible and wide application tokeniser trained on the same domain - encyclopedia data \citep{devlin-etal-2019-bert} . The Wikitext103 training data was split into 2 non-overlapping dataset of short sentences ($5-20$ tokens) and longer sentences ($21-50$ tokens). These partitions together contain $92.84\%$ of the sentence length distribution and allows for analyses of different sentence length distributions. For comparability and computational ease, a sample of $500$K was taken from both short and long sections. Thereafter, the data was partitioned into train/test as a $90\%/10\%$ with this train split further into train/validation $90\%/10\%$.

\begin{table}[!ht]
\centering
\begin{tabular}{lccc}
\hline 
& & \multicolumn{2}{c}{\textbf{Lengths}} \\ \cline{3-4}
& \textbf{Size} & \textbf{Short} & \textbf{Long}  \\ 
\hline
Train & 405K & 5-20 & 21-50\\
Validation & 45K & 5-20 & 21-50 \\
Test  & 50K & 5-20 & 21-50 \\
\hline
\end{tabular}
\caption{Wikitext103 data partitions, size and lengths.}
\label{Table:wikihowLengths}
\end{table}

\begin{figure}
\centering
\begin{minipage}{.5\textwidth}
  \centering
  \includegraphics[width=0.9\linewidth]{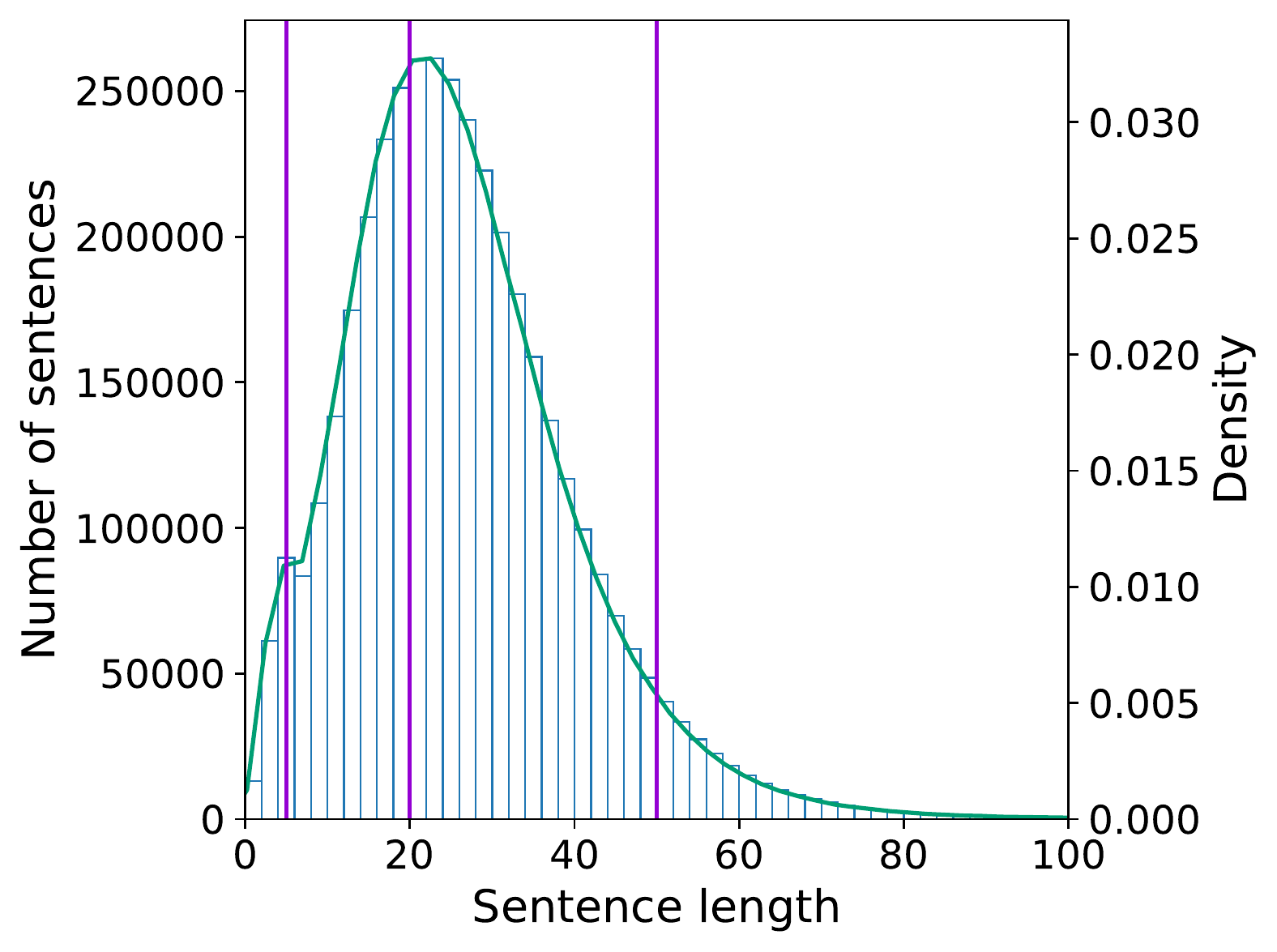}
\end{minipage}%
\begin{minipage}{.5\textwidth}
  \centering
  \includegraphics[width=0.9\linewidth]{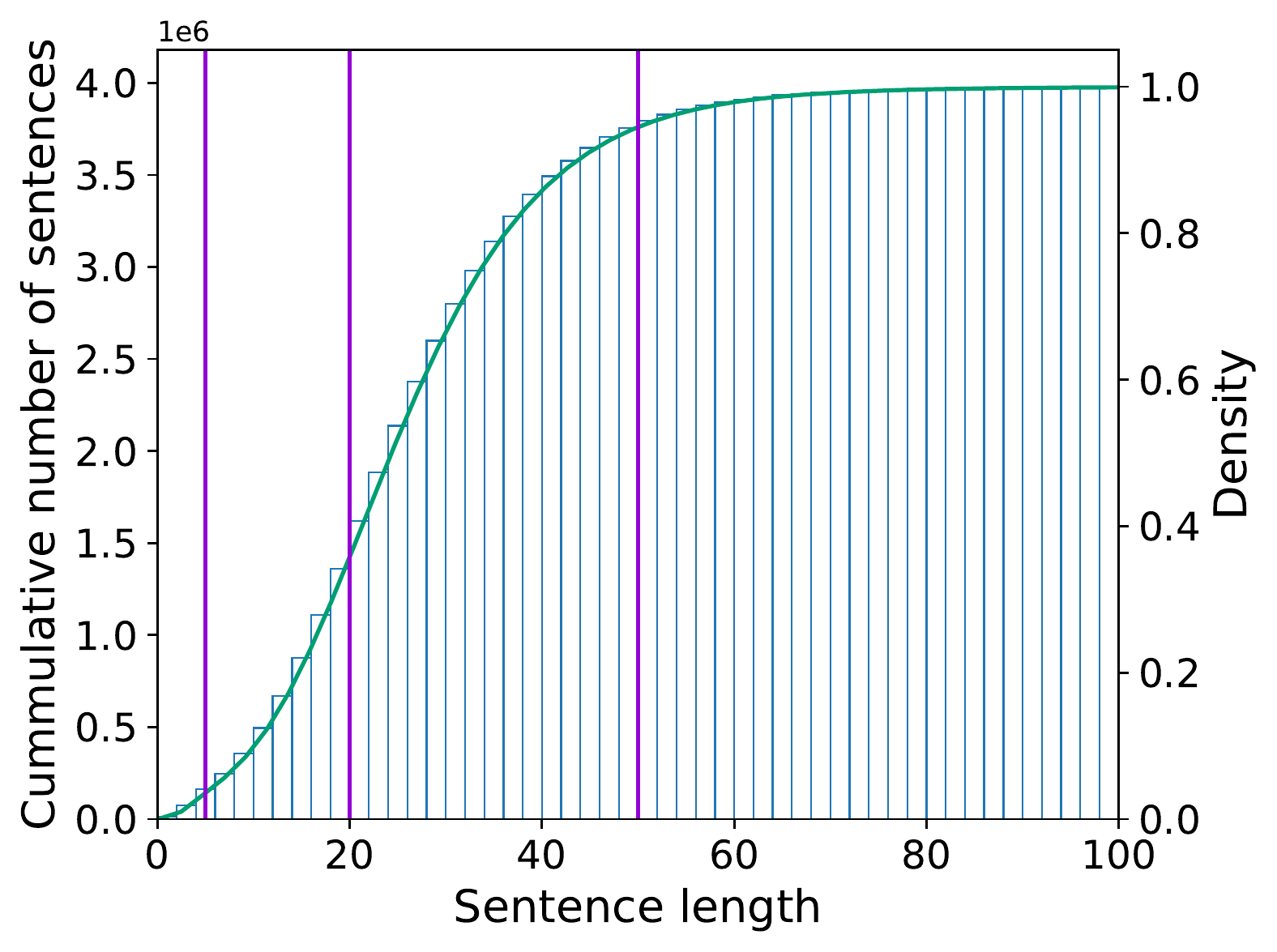}
\end{minipage}
\caption{Distribution of sentence lengths for Wikitext103 partitioned into two sections: short sentences ($5-20$ tokens) and long sentences ($20-50$ tokens) together containing $92.84\%$ of all sentence lengths.}
  \label{fig:test}
\end{figure}

\subsubsection{Training details}

All models are trained from a random initialisation, without pretraining.  For comparability, they all us use the same encoder and decoder configuration, which is a single layer Transformer encoder and decoder, each with a single attention-head. The size for the word embedding vectors, model feed forward dimension and latent projections is $256$, which leads to models of approximately $17$ million trainable parameters. The BERT base-uncased tokeniser is used for tokenisation with a vocabulary of approximately $30$K. During training we use: a constant learning rate of $5e^{-5}$, Adam optimiser \citep{kingma15_adam}, a batch size of $256$, gradient norm clipping of $0.1$ and trained for $50$ epochs (approximately $80$K steps). The loss function used is cross-entropy. As a form of regularisation we use a dropout rate of $0.1$ and the VIB parameters $\lambda^\prime_G$ and $\lambda^\prime_D$, selected through a hyperparameter search over log descending values. No annealing strategies or learning rate schedulers are used. Each model experiment takes approximately 2hrs to on NVIDIA GeForce RTX 3090.

\subsection{Baseline Models}

In addition to our NVAE model, defined in Section~\ref{sec:nvae}, we evaluate several comparable baseline models.  These baselines are all Transformer-based autoencoders, and only differ from the NVAE model in the latent representation between the encoder and decoder.  To regularise the content of vectors and provide priors for generation, we use a single-vector Gaussian VIB inserted in between the encoder and decoder, either applied to a single pooled vector or to each Transformer embedding vector independently.  We also consider alternative baselines which constrain the number of vectors.

\paragraph{Transformer (T)}
A vanilla Transformer encoder-decoder without any VIB \citep{Vaswani2017}. This model uses all the vectors and is unable to generate because there is no prior to sample from.  However, it is a reference point for the sentence reconstruction task. 

\paragraph{Variational Transformer (VT)}
A vanilla Transformer encoder and decoder with a single-vector VIB layer applied to each embedding vector independently.  Each vector at the output of the encoder is projected to $\vf{\mu}^q_i \in \mathbb{R}^{1 \times d}$ and $\text{log}(\vf{\sigma})^q_i \in \mathbb{R}^{1 \times d}$ to specify a Gaussian distribution for that vector.  The VIB layer is trained with a Gaussian $\KL$ loss function \citep{kingma2014autoencoding} which regularises the information conveyed by this vector.   This baseline uses all the vectors and is a generative model.

\paragraph{Variational Transformer Pooled (VTP)}
A version of the Variational Transformer where the vectors output by the encoder are first pooled into a single vector across the sentence length dimension.  The Gaussian VIB layer and $\KL$ loss function, as in VT above, are then applied to this one vector. Pooling methods considered are \textit{mean}, \textit{max} and just selecting the first vector, similar to the \textit{cls} \citep{devlin-etal-2019-bert}. This baseline uses a single vector and is a generative model.

\paragraph{Variational Transformer Stride (VTS)}
A version of the Variational Transformer with a location-based masking of latent vectors.  This is a hand-coded approach to reducing the number of vectors in the latent representation, where the retained vectors are spaced evenly over the input sequence.  The stride parameter $S=0.5$ masks every second vector in the latent space and stride $S=0.25$ masks every fourth vector. The Gaussian VIB layer and $\KL$ loss function are applied to each of the retained vectors, as in VT above. This baseline uses vectors for $1{-}S$ proportion of the input tokens and is a generative model.

\subsection{Reconstruction versus Generation Trade-off}

The following section addresses the following hypothesis: \textbf{H1} the NVAE Transformer model is a viable VAE as it is able to both reconstruct sentences and generate sentences by sampling from the prior.  It should be able to reconstruct sentences it was not trained on, and sampling from the prior over the latent space should generate the distribution over sentences.

We compare our model against the variational Transformer alternatives and compare using automatic reconstruction and generation metrics for language, defined in Section \ref{sec:metrics}. For computational efficiency and as a proof of concept, we run the experiments on the short sentence length partition of our data.  For consistency of terminology, and because in these experiments the NVAE samples one vector per mixture component, we use the term "vectors" to refer to both vectors output by baseline encoders and mixture components output by NVAE encoders.

\begin{table}[!ht]
\centering
\begin{tabular}{llllllll}
\hline 
               & & & & \multicolumn{2}{c}{\textbf{Reconstruction}} & \multicolumn{2}{c}{\textbf{Generation}}  \\ \cline{5-8}
\textbf{Model} & & & $\boldsymbol{\nu}$ &  \textbf{BLEU} $\uparrow$ & \textbf{PPL} $\downarrow$ & \textbf{F-PPL} $\downarrow$ & \textbf{R-PPL} $\downarrow$ \\ \hline
T & & & 1 &  99.54 \textcolor{grey}{$\pm$0.02} & 1.00 \textcolor{grey}{$\pm$0.00} & - & - \\
VT & $\lambda^\prime_G=0$ & & 1 & 99.55 \textcolor{grey}{$\pm$0.00} & 1.00 \textcolor{grey}{$\pm$0.00} & 1.33 \textcolor{grey}{$\pm$0.17} & 28.56 \textcolor{grey}{$\pm$44.29} \\
NVAE & $\lambda^\prime_G=0$ & $\lambda^\prime_D=0$ & 1 & 99.55 \textcolor{grey}{$\pm$0.00} & 1.00 \textcolor{grey}{$\pm$0.00} & 2.02 \textcolor{grey}{$\pm$0.96} & 51.25 \textcolor{grey}{$\pm$45.57} \\
\hline
VT  & $\lambda^\prime_G=0.0001$ & & 1 & 99.55 \textcolor{grey}{$\pm$0.00} & 1.00 \textcolor{grey}{$\pm$0.00} & 1.04 \textcolor{grey}{$\pm$0.01} & 1.29 \ \ \textcolor{grey}{$\pm$0.08} \\
VTP & $\lambda^\prime_G=0.0001$ & $mean$ & 0.03* & 68.28 \textcolor{grey}{$\pm$0.29} & 1.69 \textcolor{grey}{$\pm$0.01} & 2.73 \textcolor{grey}{$\pm$0.49} & 1.05  \ \ \textcolor{grey}{$\pm$0.01} \\
VTP & $\lambda^\prime_G=0.0001$ & $max$ & 0.03* & 66.17 \textcolor{grey}{$\pm$0.57} & 2.11 \textcolor{grey}{$\pm$0.06} & 1.03 \textcolor{grey}{$\pm$0.01} & 1.10  \ \ \textcolor{grey}{$\pm$0.02} \\
VTP & $\lambda^\prime_G=0.0001$ & $cls$ & 0.03* & 48.76 \textcolor{grey}{$\pm$0.33} & 4.09 \textcolor{grey}{$\pm$0.04} & 1.01  \textcolor{grey}{$\pm$0.01} & 1.28  \ \  \textcolor{grey}{$\pm$0.03}\\
VTS & $\lambda^\prime_G=0.0001$ & $S=0.25$ & 0.75 & 99.55 \textcolor{grey}{$\pm$0.00} & 1.00 \textcolor{grey}{$\pm$0.00} & 1.01 \textcolor{grey}{$\pm$0.01} & 1.30 \ \ \textcolor{grey}{$\pm$0.06} \\
VTS & $\lambda^\prime_G=0.0001$ & $S=0.5$ & 0.5 & 99.46 \textcolor{grey}{$\pm$0.04} & 1.00 \textcolor{grey}{$\pm$0.00} & 1.13 \textcolor{grey}{$\pm$0.12} & 7.19  \ \ \textcolor{grey}{$\pm$6.91} \\
VTS & $\lambda^\prime_G=0.0001$ & $S=0.75$ & 0.25 & 95.32 \textcolor{grey}{$\pm$0.14} & 1.08 \textcolor{grey}{$\pm$0.01} & 1.04 \textcolor{grey}{$\pm$0.13} & 1.62  \ \ \textcolor{grey}{$\pm$0.34} \\
VTS & $\lambda^\prime_G=0.0001$ & $S=0.9$ & 0.1 & 70.40 \textcolor{grey}{$\pm$0.32} & 1.73 \textcolor{grey}{$\pm$0.01} & 1.03\textcolor{grey}{$\pm$0.02} & 1.32  \ \ \textcolor{grey}{$\pm$0.07} \\
\hline
NVAE & $\lambda^\prime_G=0.001$ &  $\lambda^\prime_D=1$  & 0.79 & 99.46 \textcolor{grey}{$\pm$0.07} & 1.00 \textcolor{grey}{$\pm$0.00} & 1.01 \textcolor{grey}{$\pm$0.01} & 2.87  \ \ \textcolor{grey}{$\pm$1.39} \\
NVAE & $\lambda^\prime_G=0.01$ &  $\lambda^\prime_D=1$  & 0.47 & 92.26 \textcolor{grey}{$\pm$4.90} & 1.18 \textcolor{grey}{$\pm$0.13} & 1.00 \textcolor{grey}{$\pm$0.00} & 5.06  \ \ \textcolor{grey}{$\pm$1.90} \\
NVAE & $\lambda^\prime_G=0.1$ &  $\lambda^\prime_D=1$  & 0.36 & 55.38 \textcolor{grey}{$\pm$14.25} & 3.25 \textcolor{grey}{$\pm$1.22} & 1.00 \textcolor{grey}{$\pm$0.00} & 3.38 \ \  \textcolor{grey}{$\pm$0.98} \\
NVAE & $\lambda^\prime_G=1$ &  $\lambda^\prime_D=1$  & 0.17 & 31.94 \textcolor{grey}{$\pm$2.76} & 7.91 \textcolor{grey}{$\pm$2.41} & 1.00 \textcolor{grey}{$\pm$0.00} & 1.29  \ \ \textcolor{grey}{$\pm$0.04} \\
\hline
\end{tabular}
\caption{Results for regularisation and generation on validation Wikitext103 (short partition) averaged over 5 seeds. Reconstruction metrics are validation perplexity ({PPL}) and {BLEU}. Sample quality metrics are forward perplexity ({F-PPL}) and reverse perplexity ({R-PPL}). The average proportion of latent vectors retained during evaluation is reported by $\boldsymbol{\nu}$. ~ *The VTP models only use a single vector.}
\label{tab:generationReconstuction}
\end{table}

The NVAE model uses two regularisation parameters $\lambda^\prime_G$ and $\lambda^\prime_D$ to regularise the content and number of vectors in the latent space. When unregularised -  regularisation parameters are zeroed, as in the third line of Table \ref{tab:generationReconstuction} - the optimal model reverts to a Transformer (T) as expected. As such, it provides the best and worse case reconstruction and generation metrics respectively. 

The baseline VT regularises only the content of the latent space and when regularised achieves the best case reconstruction score with good generation metrics, although the F-PPL is not the lowest, as can also be seen in the sampled text in Table~\ref{tab:SamplesShort}. Due to the simplistic nature of the task we conjecture that a more compressed latent space may exist, represented by fewer vectors. The most compressed latent space would be a single vector latent space, such as the VTP variants. These models do not have enough capacity to both reconstruct well (high BLEU) and generate fluent (low F-PPL) and diverse, plausible sentences (low R-PPL). The results in Table~\ref{tab:generationReconstuction} show that the VTP models tend to achieve poor reconstruction scores, but generate well. Upon further inspection, this is attained by generating long, diverse sentences as seen in Table~\ref{tab:SamplesShort}.

The VTS models are hand-coded solutions to providing an intermediate level of compression between keeping all the vectors and dropping all but one.  These models demonstrate that there exist models that are able to both reconstruct and generate effectively with as few as a quarter or half the number of original vectors, as seen in Table~\ref{tab:generationReconstuction}. 

The NVAE models are able to learn to dynamically determine the level of compression based on the information content of the input text, including simultaneously regularising both the content and quantity of latent vectors. 
Table~\ref{tab:generationReconstuction} shows that there are NVAE model settings where they are able to both reconstruct and generate, including consistently good F-PPL, also reflected in the samples of generated text shown in Table~\ref{tab:SamplesShort}.  But these initial experiments did not find hyper-parameters that on average over 5 seeds balanced the trade off between reconstruction and generation as well as the VTS models, with greater variance in R-PPL.  We suspect this variance is a result of the more challenging optimisation process caused by the need to learn to balance the quantity of vectors and information they carry.

\begin{figure}[!ht]
\centering
\begin{minipage}{.5\textwidth}
  \centering
   \includegraphics[width=\linewidth]{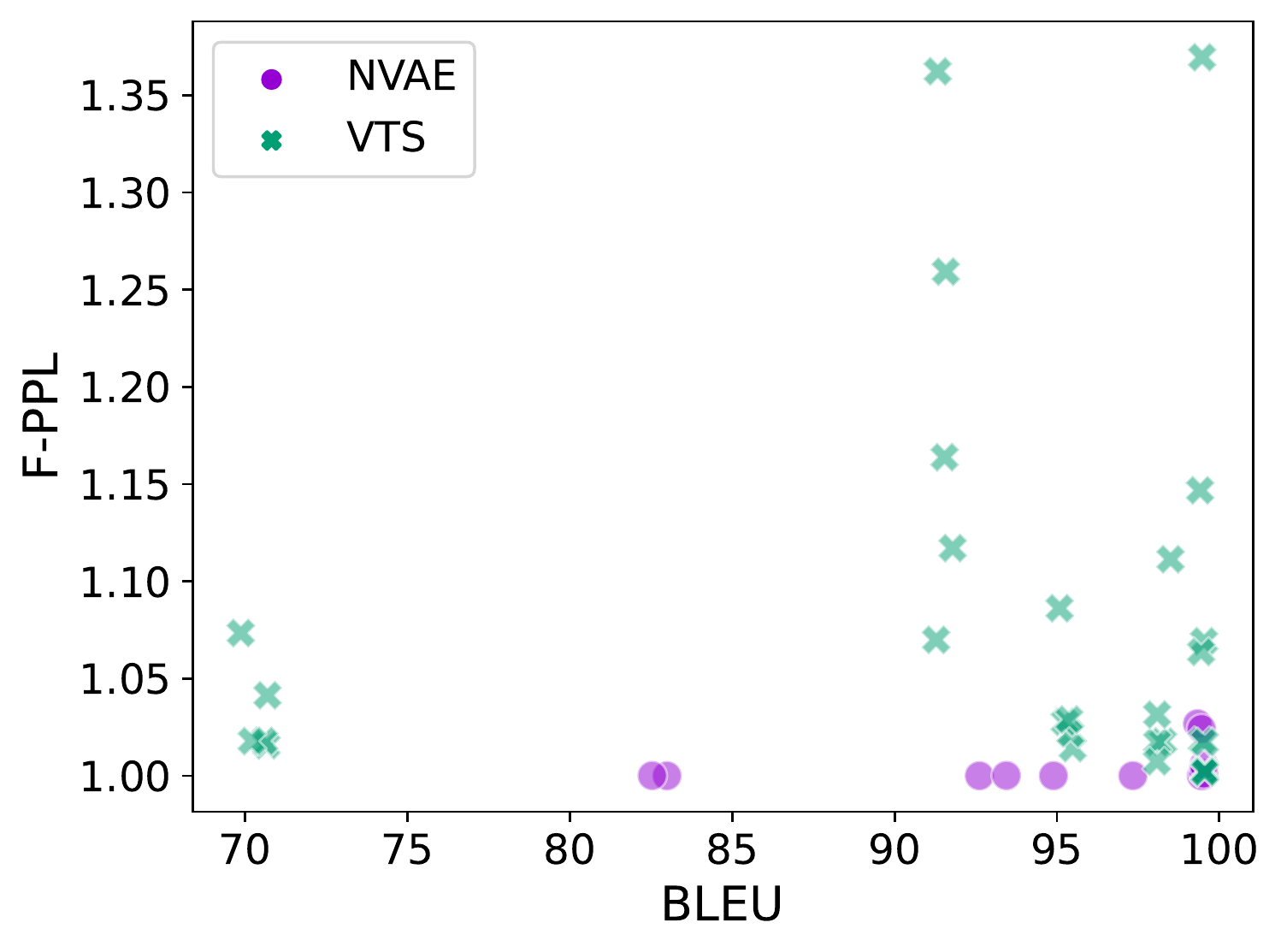}
\end{minipage}%
\begin{minipage}{.5\textwidth}
  \centering
   \includegraphics[width=\linewidth]{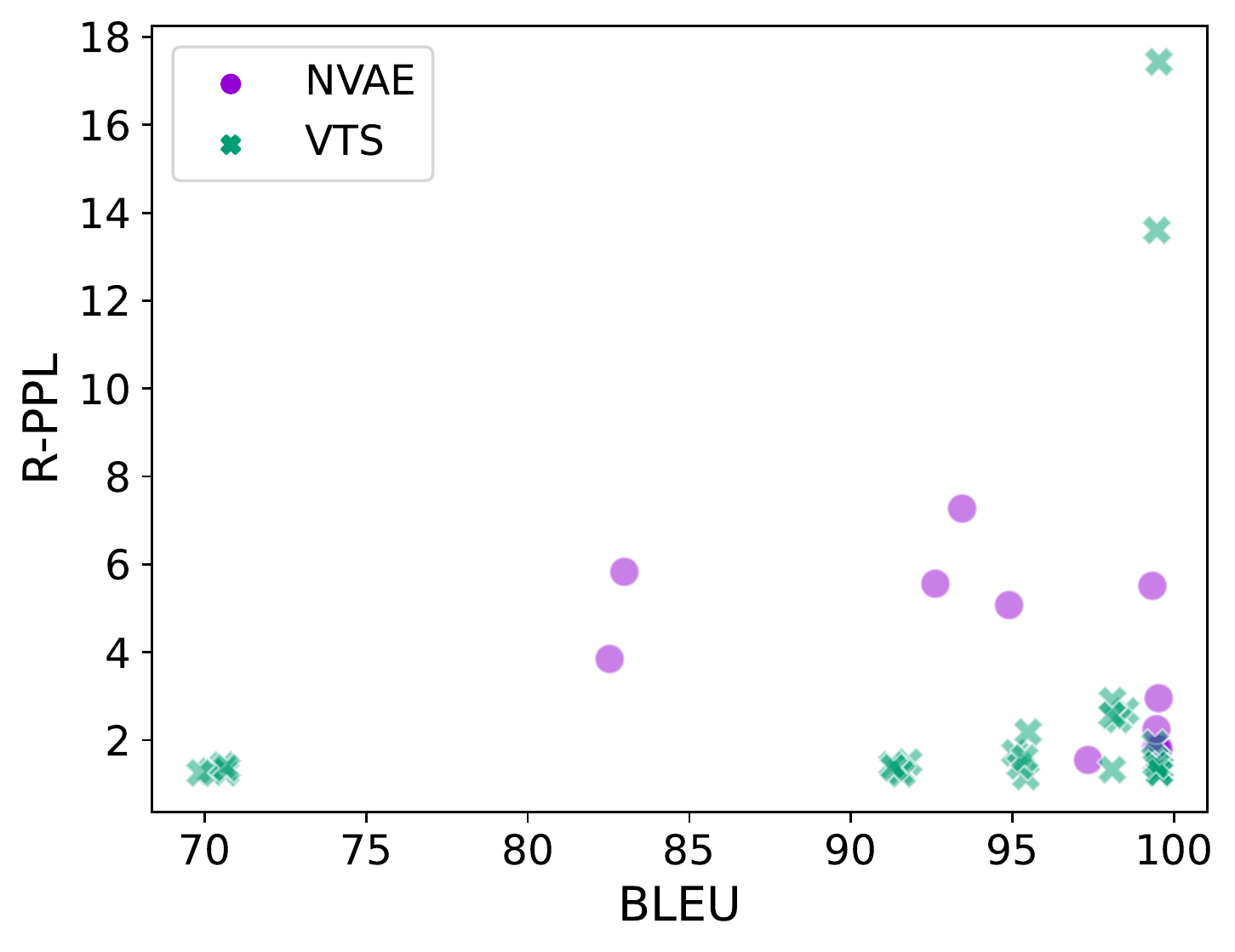}
\end{minipage}
\caption{Reconstruction BLEU and generation F-PPL (left) and R-PPL (right) for validation Wikitext103 (short partition) data across hyperparameters and seeds. Lower generation F-PPL, R-PPL and higher reconstruction BLEU are better models.}
\label{fig:bleu-rppl-short}
\end{figure}

To investigate this variance, Figure~\ref{fig:bleu-rppl-short} shows the generation-regularisation trade off across seeds and hyperparameters for the NVAE and VTS models. This plot confirms that there do exist NVAE models that are able to score both a high reconstruction BLEU and a low generation F-PPL and R-PPL. Two VTS $S=0.5$ models by chance generated poor samples which resulted in the large R-PPL scores and variance seen in Table \ref{tab:generationReconstuction}.

In this section we showed that the NVAE Transformer model is a viable VAE which is able to reconstruct and generate natural language. In contrast to the VTS, the NVAE has more variance across seeds. However, the NVAE is able to learn to dynamically select the the content and number of vectors based on the information in the text, without hand-coding the proportion of vectors required per number of tokens.  The next section will further analyse these regularisation abilities of NVAE.

\begin{table}[!ht]
\centering
\begin{tabular}{l|p{0.8\linewidth}}
\hline 
& \textbf{Samples Wikitext103 (short partition)} \\ 
\hline
\textbf{Original Text} & 
\tabitem  we never worked together in terminator 2, but we're in that movie together. \\
& \tabitem objects such as trees were added to complete the design. \\
& \tabitem in 1912, he won a rhodes scholarship to the university of oxford in england. \\
& \tabitem support for birmingham city ran in the family his uncle played for the club in the 1920s.\\
& \tabitem thou, silent form, dost tease us out of thought \\
\hline

\textbf{VT} & 
\tabitem  the lionel marriage shawn subtropical [UNK] in cal cal to and november.. \\
$\lambda^\prime_G = 0.0001$ & \tabitem the wings abroad navy,, the the producing abroad navy, countries.. \\
 & \tabitem gene disc valentine. in mayw songwriterdestin. in.. fin 98tin. \\
& \tabitem china in plot, struggled fragment genus [UNK] may may draft weakened creek genus in may plotrst.\\
& \tabitem our first theirde immaculate round was first first first their 2006 2006 \\
\hline

\textbf{VTP}
 & \tabitem henry nor were evenidness, member in formal estate inxi enacted \\
$\lambda^\prime_G= 0.0001$ & \tabitem you are considering her biblical style, her entrato wanted eroting [UNK] 2009 \\
$max$ & \tabitem the couple notes more in da is illic of more inineow \\
& \tabitem morris the improved is undertaken,real his directionorg machine gun residence point \\
& \tabitem administration of afghanistan pondered any member to the singles being purposely flourishing planned singles from northeastward. \\
\hline

\textbf{VTS} 
 & \tabitem 41st party include the stuffed and an original tests \\
$\lambda^\prime_G = 0.0001$ & \tabitem football of the 1933, arages of the expedition \\
$S = 0.75$ & \tabitem grace, the multimedia, and the performance, was 110. \\
& \tabitem its the20s to the most powerful lodgefe.\\
& \tabitem this, the misery that nytokoto this way. \\
\hline

\textbf{NVAE} 
& \tabitem and is the session in bergerome used on a. u. march and the 2010. time \\
$\lambda^\prime_G = 0.01$ & \tabitem port were spent a mistaken to the and the show of italy or from its time. 4. \\
$\lambda^\prime_D = 1$ & \tabitem the second travel with was much more and the request of 2006 and pen,. 2010. no \\
& \tabitem the game, might orchestralnta and had from the game of conservation, behind.. 2010. \\
& \tabitem the weather canson paul spend three and was actually way some attack and from the 2010.. \\
\hline

\end{tabular}
\caption{The first 5 samples drawn from the best models (lowest generation scores in-distribution) trained on Wikitext103 (short partition). Samples drawn using the empirical length distribution of Wikitext103 (short partition) for $\kappa_0$.}
\label{tab:SamplesShort}
\end{table}

\subsection{Regularisation Analysis}
\label{sec:regularisation_analysis}

This section addresses the hypothesis: \textbf{H2} the NVIB layer is able to dynamically regularise both the number and informativeness of component distributions in the latent space.  We regard the informativeness of components to be well regularised if the generated samples drawn from the prior distribution achieve low F-PPL and R-PPL scores. The number of vectors is reported as the average proportion  $\boldsymbol{\nu}$ of latent vectors retained during evaluation.

\begin{figure}[!ht]
\centering
\begin{minipage}{.495\textwidth}
  \centering
  \includegraphics[width=\linewidth]{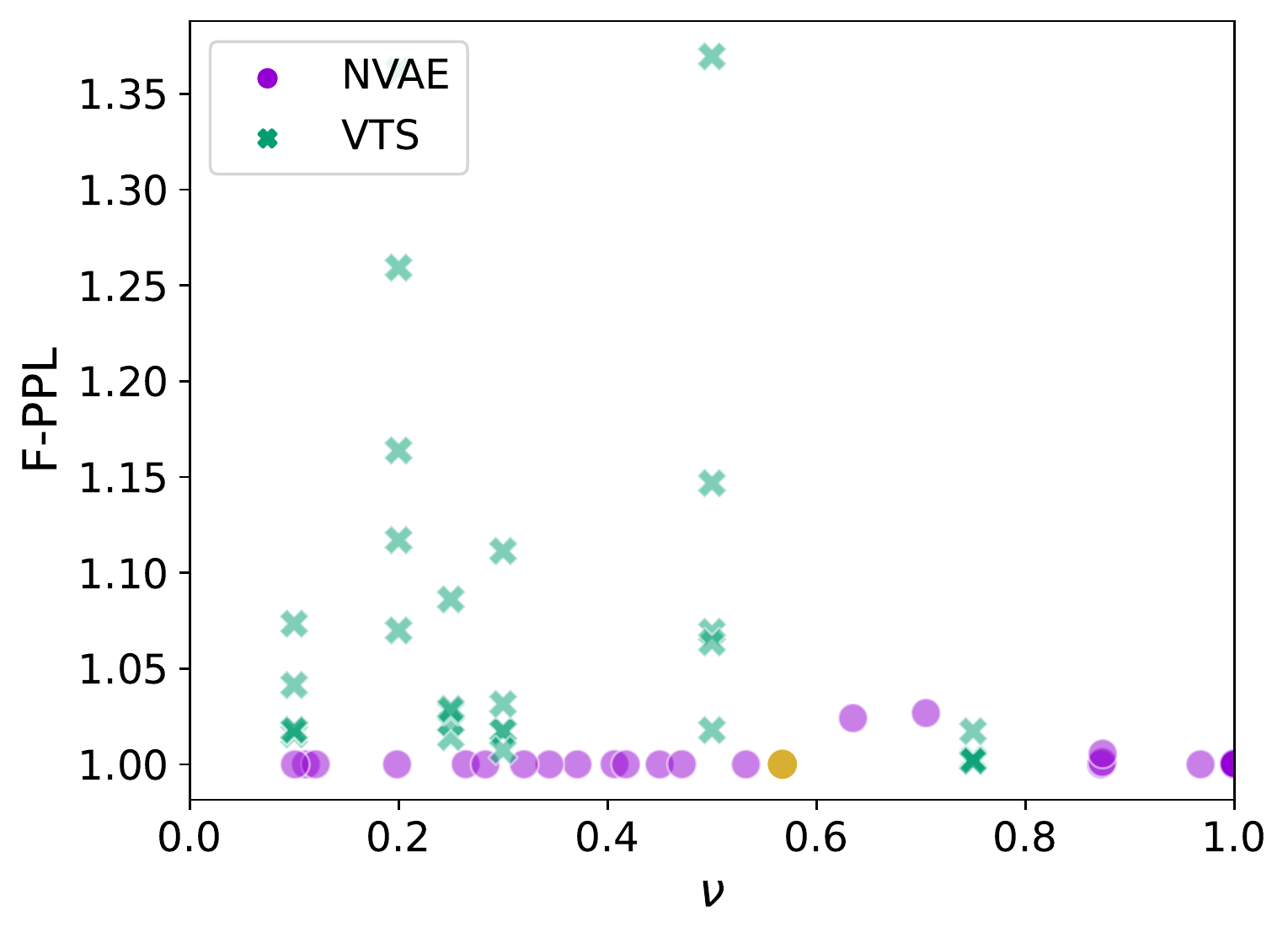}
\end{minipage}
\begin{minipage}{.495\textwidth}
  \centering
  \includegraphics[width=\linewidth]{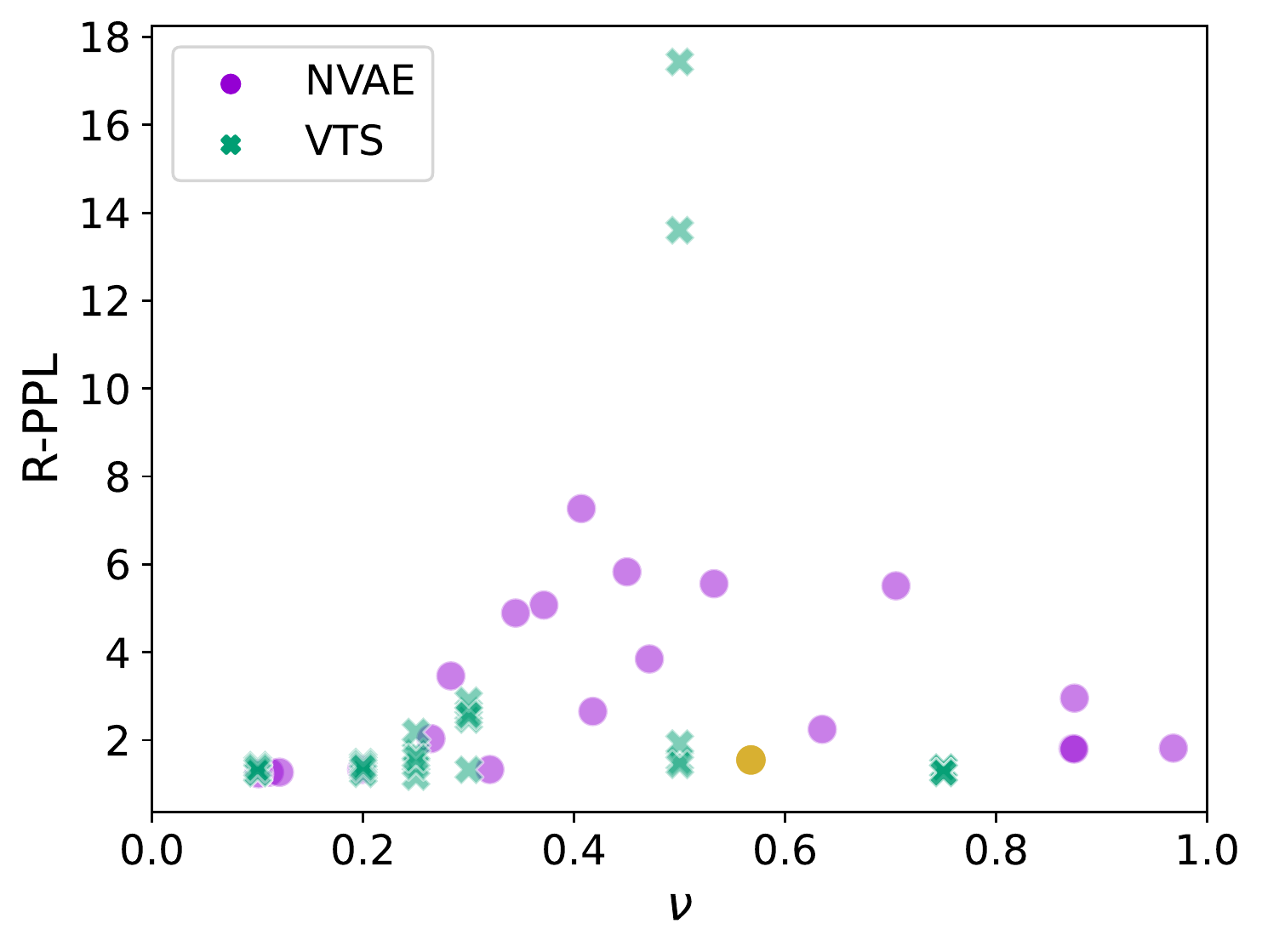}
\end{minipage}
\caption{Generation F-PPL (left) and R-PPL (right) vs the average proportion of latent vectors retained during evaluation $\boldsymbol{\nu}$.}
\label{fig:fppl-rppl-vectors}
\end{figure}

Figure \ref{fig:fppl-rppl-vectors} displays the trade off of the average proportion of latent vectors $\boldsymbol{\nu}$ against the automated generation metrics, across seeds and hyperparameters. In general, the NVAE models have better performance in F-PPL (left) than R-PPL (right) which suggests that these samples are fluent but might lack diversity. However, there exists a NVAE model that is able to achieve low F-PPL and R-PPL scores with just over half the number of vectors, shown in \textcolor{gold}{gold}.

\begin{figure}[!ht]
\centering
\begin{minipage}{.495\textwidth}
  \centering
  \includegraphics[width=\linewidth]{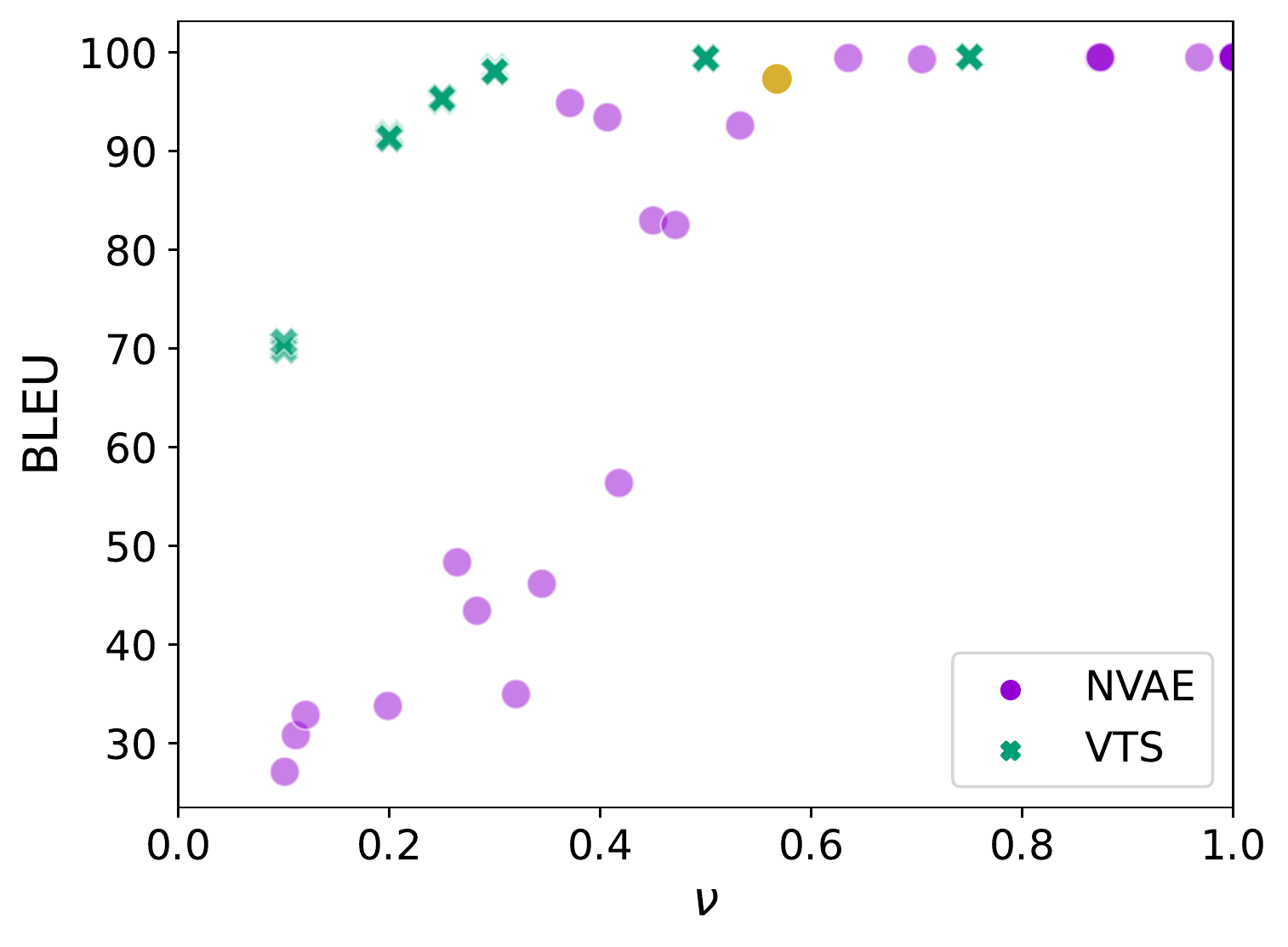}
\end{minipage}
\begin{minipage}{.495\textwidth}
  \centering
  \includegraphics[width=\linewidth]{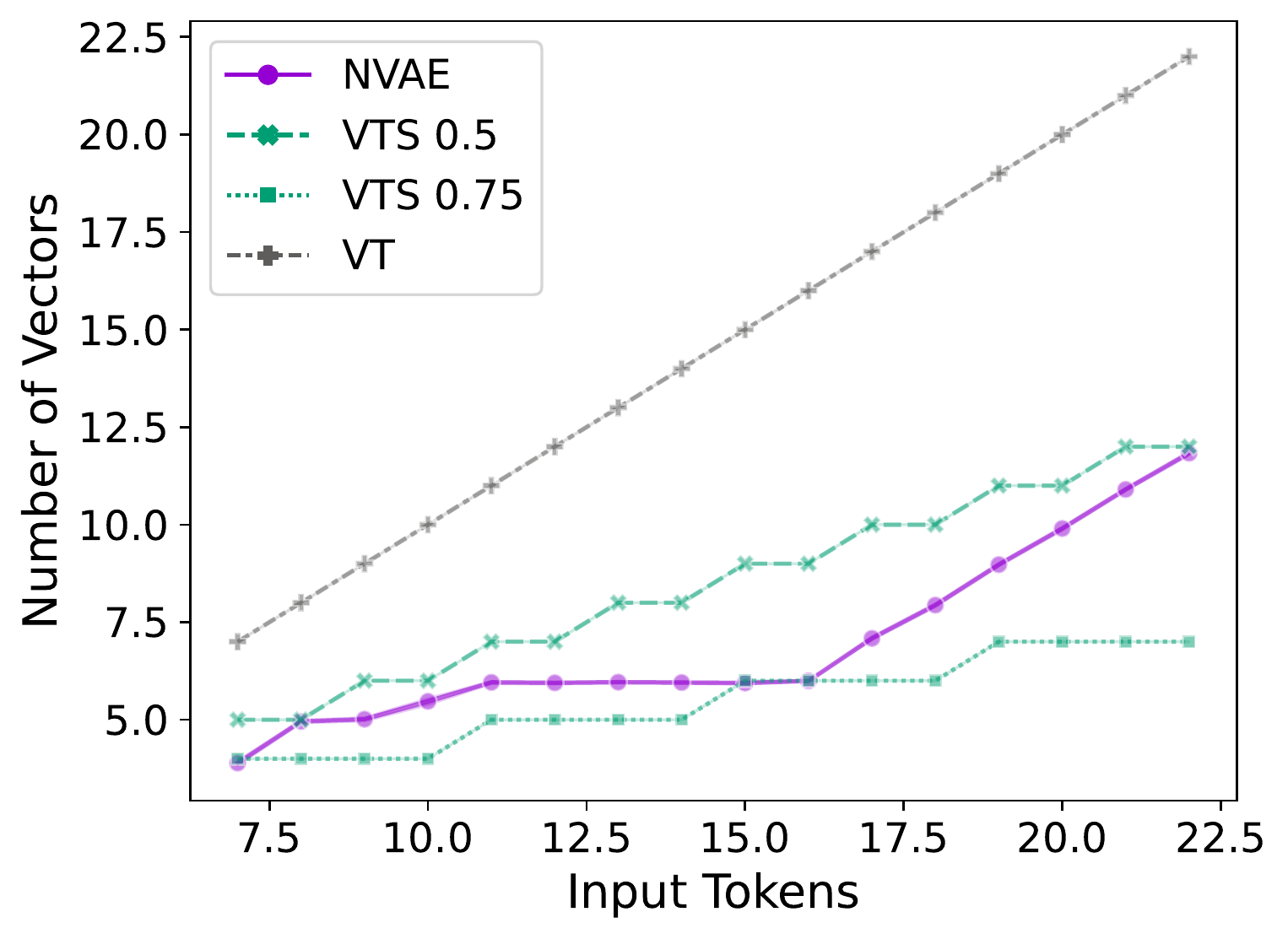}
\end{minipage}
\caption{Reconstruction BLEU vs the average proportion of vectors retained during evaluation $\boldsymbol{\nu}$ (left). Number of latent vectors retained during evaluation over the number of input tokens for a NVAE model (right).}
\label{fig:bleu-seq-vectors}
\end{figure}

Figure \ref{fig:bleu-seq-vectors} (left) shows the relationship between reconstruction BLEU score and $\boldsymbol{\nu}$. There is a logistic upward trend which shows that more vectors results in better reconstruction. We see that there exists a space of models with roughly half the number of vectors that is able to reconstruct near perfectly. The same model described before is highlighted in \textcolor{gold}{gold}. This model is able to regularise the content of vectors as it is able to generate as effectively as VT models.

Considering the NVAE model that is able to balance both reconstruction and generation, Figure \ref{fig:bleu-seq-vectors} (right) plots the average number of active latent vectors during evaluation as a function of the number of input tokens. This shows that the NVAE model dynamically adjusts the number of latent vectors required for different number of input tokens. 

\begin{figure}[!ht]
\centering
  \includegraphics[width=0.5\linewidth]{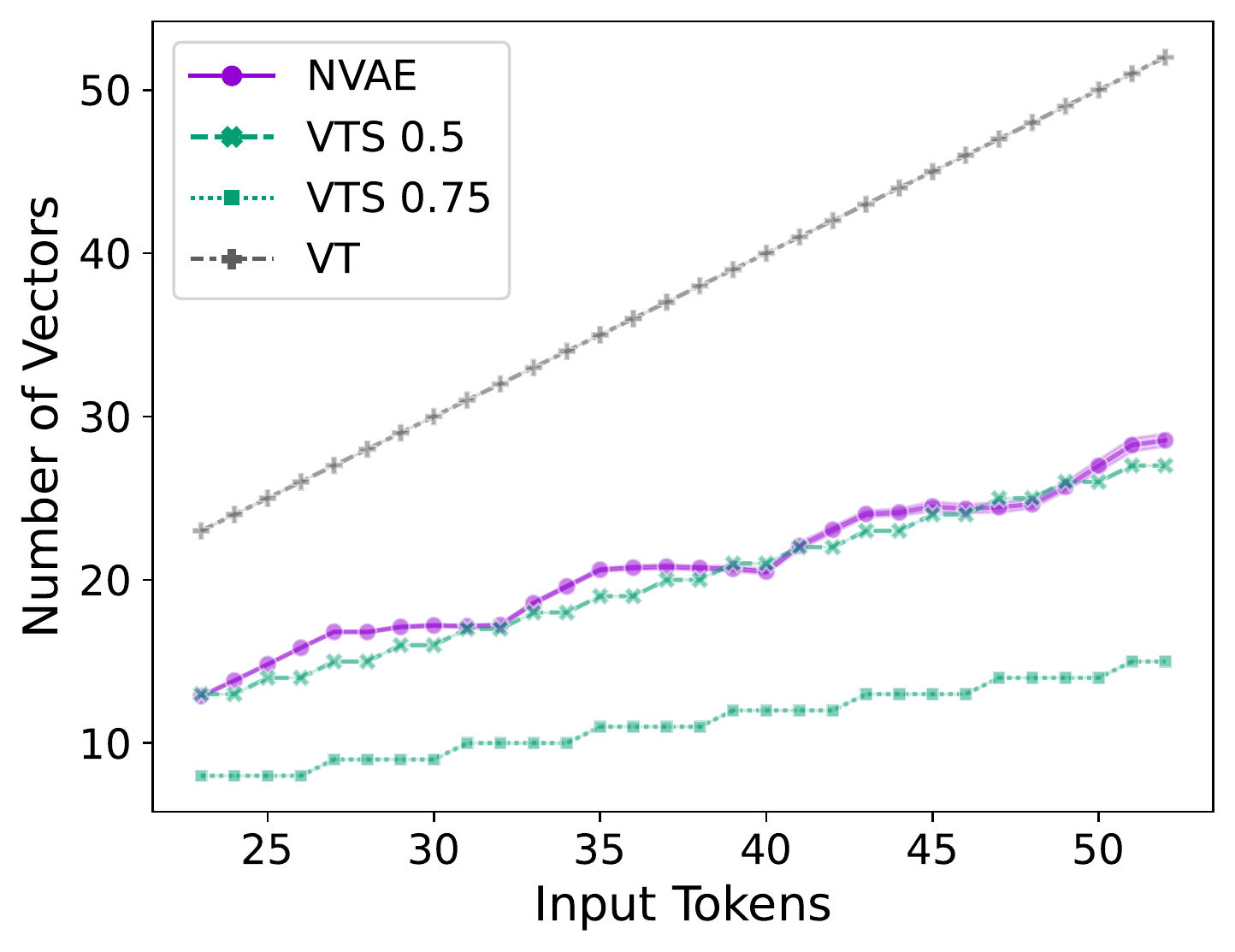}
\caption{Number of latent vectors retained during evaluation over the number of input tokens in Wikitext103 (long partition) for the NVAE model.}
\label{fig:number_vectors_OOD}
\end{figure}

Using the Wikitext (long partition) we are able to evaluate how the model is able to generalise to sentence lengths outside of the training distribution. Figure \ref{fig:number_vectors_OOD} displays the same model's quantity of latent vectors used during out-of-distribution text reconstruction.  The NVAE has generalised to using nearly the exact same number of vectors as the hand-coded VTS with $S=0.5$ for these sentence lengths which it was not trained on.

In conclusion, we have shown that there exist NVAE models that are able to dynamically regularise both quantity and content of vectors in the latent space. The sentence length generalisation hints that the NVAE is able to generalise beyond the trained distribution due to the prior distribution. We leave further analysis to future work.

\section{Discussion}

In this paper we focus on developing an effective variational information bottleneck for regularising Transformer embeddings, which our initial empirical evaluations confirm.  However, we believe that the nonparametric Bayesian framework proposed here has a wider significance in the study of natural language, and cognitive science.  The use of mixture distributions to formalise attention-based representations is extremely natural, fundamentally because they are both permutation invariant, which implies exchangeability is a key property for Bayesian approaches to attention-based representations.  The unprecedented empirical success of Transformers is thus strong evidence that exchangeability is a fundamental property of natural language.

In particular, this paper supports the argument that exchangeability is the fundamental mechanism which allows natural language understanding to extend to arbitrarily long texts, 
just like it is a fundamental mechanism which allows Bayesian methods to extend to nonparametric distributions \citep{Jordan10}.  Given the argument of \citet{henderson-2020-unstoppable} that Transformers represent a continuation of the Computational Linguistic tradition of unbounded structured representations, this variational Bayesian framework represents a point of convergence between this tradition and the probabilistic models of Cognitive Science and Neuroscience.

\section{Conclusions}

In this work, we formalise the latent representations of attention-based models as mixture distributions over a vector space, in Section \ref{sec:framework}, and propose a nonparametric variational information bottleneck (NVIB) to regularise these latent representations, in Section \ref{sec:variational}.  Using this NVIB model, we propose a nonparametric variational autoencoder (NVAE), in Section \ref{sec:nvae}, which uses a Transformer encoder to embed text in a nonparametric space of distributions over mixture distributions, and uses a Transformer decoder to generate text given a sample from this latent space.  In accordance with our Bayesian framework for attention-based representations, the decoder uses a generalisation of cross attention as query denoising to access the latent representation.  At test time, the latent representation is the mean of the sampled discrete distributions, which is a continuous mixture of Gaussians.

This nonparametric Bayesian formalisation of attention-based representations captures two key properties of attention, namely that the vectors in the input set-of-vectors are permutation invariant, and that the set can vary widely in size.  Our NVIB model adds to these properties the ability to regularise attention-based representations so that the size of the representation is appropriate for the complexity of the input being encoded.  This is a crucial ability for encoding text, where the size of a text can vary enormously.  

Initial results, in Section \ref{sec:evaluation}, indicate that this model does regularise the induced latent spaces as desired, and is a viable VAE in that it is able to both reconstruct input sentences and generate a good distribution over sentences from the prior.
Future work will increase the scale of these experiments, address the usefulness of this induction method for NLP tasks, and investigate the wider significance of this nonparametric variational Bayesian approach to modelling natural language and the thoughts it expresses.

\section{Limitations}

\paragraph{Theory}
We have explored a relatively efficient version of sampling from the posterior DP, with one vector sampled from each mixture component.  The Bayesian framework suggests alternatives, which we intend to explore in future work.

\paragraph{Implementation}
The NVIB takes advantage of two sources of variational noise, which naturally makes the model more challenging to train. This is a known problem with VAEs. The two parts of the $\KL$ divergence are not easily decoupled for easy tuning of both the content and quantity of vectors.
Also, we have only experimented with single-head attention.  Multi-head attention is compatible with the theoretical model, but more complicated for parallelisable implementation.

\paragraph{Experimentation}
The current experiments are a proof of concept and show potential for future work. To limit the computational requirements, small data sizes and short sentence lengths were used and pretraining was excluded. The generation metrics to evaluate the quality of diversity and fluency are not necessarily representative and a human evaluation should be considered in future work.

\section*{Acknowledgements}

We would like to thank Florian Mai, Andrei Catalin Coman, Melika Behjati, Andreas Marfurt, and other members of the Idiap NLU group for helpful comments and suggestions.  This work was funded in part by the Swiss National Science Foundation under the NCCR grant Evolving Language, Swiss National Science Foundation Agreement \#51NF40\_180888.

\bibliographystyle{acl_natbib}
\bibliography{anthology,mybib,ff_references}

\newpage
\setcounter{section}{0}
\renewcommand\thesection{\Alph{section}}
\renewcommand\thesubsection{\thesection.\Alph{subsection}}

\section{Deriving the Factorised Dirichlet Process}
\label{sec:fdpderivaion}

For notational convenience, in this section we use $c$ as the number of components for the base distribution instead of $c+1$.  This is still intended to include both the output of the encoder and the prior component.

Here we provide the proof that
\begin{eqnarray*}
  \FDP(\vf{G}^q,\vf{\alpha}^q)
  &=& \DP(G^q_0,\alpha^q_0)
\end{eqnarray*}
where $\vf{G}^q=({G}^q_1,\ldots,{G}^q_{c})$, $\vf{\alpha}^q=({\alpha}^q_1,\ldots,{\alpha}^q_{c})$, $G^q_0 = \sum_{i=1}^{c} \frac{{\alpha}^q_i}{\alpha^q_0} {G}^q_i$, $\alpha^q_0 = \sum_{i=1}^{c} {\alpha}^q_i$, and $F\sim \FDP(\vf{G}^q,\vf{\alpha}^q)$ is defined as
\begin{eqnarray*}
  F &=& \sum_{i=1}^{c} {\rho}_i {F}_i
  \\
  \vf{\rho} &\sim& \Dir({\alpha}^q_1,\ldots,{\alpha}^q_{c})
  \\
  {F}_i &\sim& \DP({G}^q_i,{\alpha}^q_i)
  \mbox{~~for~} i=1,\ldots,c
\end{eqnarray*}

We start with the definition of a DP as an infinite symmetric Dirichlet distribution.
A Dirichlet process $F\sim \DP(G_0,\alpha_0)$ can be defined as the limit of a sequence of finite Dirichlet distributions (see \citep{Teh2010}):
\begin{eqnarray*}
  F &=& \sum_{k=1}^\infty {\pi}_k \delta_{\vf{z}_k}
  \\
  \vf{\pi} &\sim& \lim_{\kappa_0\rightarrow \infty} \Dir(\frac{\alpha_0}{\kappa_0},\overset{\kappa_0}{\ldots},\frac{\alpha_0}{\kappa_0})
  \\
  \vf{z}_k &\sim& G_0 \mbox{~~for~} k=1,\ldots,\infty
\end{eqnarray*}
Note that the weights $\vf{\pi}$ and the vectors $\vf{z}_k$ are independent of each other, so we can treat these two issues separately.

For the vectors, we know that after generating an infinite number of $\vf{z}_k$ from $G_0$, a proportion of exactly $\frac{{\alpha}^q_i}{\alpha^q_0}$ of them will be generated from ${G}^q_i$.  For a finite number of vectors $\kappa_0$, let ${\kappa}_i$ be the number of $\vf{z}_k$ generated from ${G}^q_i$, for each $i$.  So we have
\begin{eqnarray*}
\lim_{\kappa_0\rightarrow \infty}~ \frac{{\kappa}_i}{\kappa_0} &=& \frac{{\alpha}^q_i}{\alpha^q_0}
\end{eqnarray*}
Given the exchangeability of Dirichlet distributions, we can renumber the $\kappa_0$ categories of $\Dir(\frac{\alpha_0}{\kappa_0},\overset{\kappa_0}{\ldots},\frac{\alpha_0}{\kappa_0})$ so that $\vf{\pi} = ({\pi}_{11},\ldots,{\pi}_{1{\kappa}_1},~\overset{c}{\ldots},~{\pi}_{c1},\ldots,{\pi}_{c{\kappa}_c})$ and the ${\pi}_{i1},\ldots,{\pi}_{i{\kappa}_i}$ are all weights for vectors $\vf{z}_{ij}$ generated from component $i$.
\begin{eqnarray*}
  \vf{z}_{ij} &\sim& {G}^q_i \mbox{~~for~} i=1,\ldots,c,~ j=1,\ldots,{\kappa}_i
\end{eqnarray*}

For the weights, we again consider the case of finite $\kappa_0$ before taking the limit as $\kappa_0$ goes to infinity, using the above indexing where categories $ij$ are partitioned according to their vector's base distribution component $i$.  We define $({\rho}_1,\overset{c}{\ldots},{\rho}_c)$ to be the vector of total weights ${\rho}_i = \sum_{j=1}^{{\kappa}_i} {\pi}_{ij}$ for each of these partitions $i$.  By the rule for merging categories in a Dirichlet distribution, we know that these total weights are themselves distributed according to a Dirichlet distribution.
\begin{eqnarray*}
  ({\rho}_1,\overset{c}{\ldots},{\rho}_c)
  &\sim& \Dir({\alpha}^q_1,\overset{c}{\ldots},{\alpha}^q_c)
\end{eqnarray*}

Now we take advantage of the neutrality property of Dirichlet distributions.
It states that this vector $({\rho}_1,\overset{c}{\ldots},{\rho}_c)$ of partition weights and all of the vectors $(\frac{{\pi}_{i1}}{{\rho}_i},\overset{{\kappa}_i}{\ldots},\frac{{\pi}_{i{\kappa}_i}}{{\rho}_i})$ of normalised weights inside each partition are independent.  In essence, this means that the only way that the weights inside each partition constrain each other is through normalisation, so when normalisation is factored out they become independent.
This independence allows us to compute the distribution over $(\frac{{\pi}_{i1}}{{\rho}_i},\overset{{\kappa}_i}{\ldots},\frac{{\pi}_{i{\kappa}_i}}{{\rho}_i})$ by simply marginalising out all the other categories.  We first merge all the categories outside partition $i$ into a single category, whose weight is thus $1-{\rho}_i$.  This gives us the marginalised Dirichlet distribution
$({\pi}_{i1},\overset{{\kappa}_i}{\ldots},{\pi}_{i{\kappa}_i},~(1-{\rho}_i))$
$\sim~ \Dir(\frac{\alpha^q_0}{\kappa_0},\overset{{\kappa}_i}{\ldots},\frac{\alpha^q_0}{\kappa_0},~ \alpha^q_0(1-\frac{{\kappa}_i}{\kappa_0}))$.
\begin{eqnarray*}
  &&\hspace{-10ex}
  d({\pi}_{i1},\overset{{\kappa}_i}{\ldots},{\pi}_{i{\kappa}_i},~(1-{\rho}_i)\;;~
  \frac{\alpha^q_0}{\kappa_0},\overset{{\kappa}_i}{\ldots},\frac{\alpha^q_0}{\kappa_0},~ \alpha^q_0(1-\frac{{\kappa}_i}{\kappa_0}))
  \\
  &=& \frac{\Gamma(\alpha^q_0)}{\Gamma(\alpha^q_0(1-\frac{{\kappa}_i}{\kappa_0}))\prod_{j=1}^{{\kappa}_i} \Gamma(\frac{\alpha^q_0}{\kappa_0})}
  (1-{\rho}_i)^{\alpha^q_0(1-\frac{{\kappa}_i}{\kappa_0})-1}\prod_{j=1}^{{\kappa}_i} {{\pi}_{ij}}^{\frac{\alpha^q_0}{\kappa_0}-1}
  \\
  &=& \frac{\Gamma(\alpha^q_0)}{\Gamma(\alpha^q_0(1-\frac{{\kappa}_i}{\kappa_0}))\prod_{j=1}^{{\kappa}_i} \Gamma(\frac{\alpha^q_0}{\kappa_0})}
  (1-{\rho}_i)^{\alpha^q_0(1-\frac{{\kappa}_i}{\kappa_0})-1} ({\rho}_i)^{\alpha^q_0\frac{{\kappa}_i}{\kappa_0}-1}
  ~(\prod_{j=1}^{{\kappa}_i} (\frac{{\pi}_{ij}}{{\rho}_i})^{\frac{\alpha^q_0}{\kappa_0}-1})
\end{eqnarray*}
Now we can marginalise out the weight of the outside category by integrating over ${\rho}_i$.
\begin{eqnarray*}
  &&\hspace{-10ex}
  \int_{{\rho}_i=0}^1 \frac{\Gamma(\alpha^q_0)}{\Gamma(\alpha^q_0(1-\frac{{\kappa}_i}{\kappa_0}))\prod_{j=1}^{{\kappa}_i} \Gamma(\frac{\alpha^q_0}{\kappa_0})}
  (1-{\rho}_i)^{\alpha^q_0(1-\frac{{\kappa}_i}{\kappa_0})-1} ({\rho}_i)^{\alpha^q_0\frac{{\kappa}_i}{\kappa_0}-1}
  ~(\prod_{j=1}^{{\kappa}_i} (\frac{{\pi}_{ij}}{{\rho}_i})^{\frac{\alpha^q_0}{\kappa_0}-1}) ~d{\rho}_i
  \\
  &=& \left( \frac{\Gamma(\alpha^q_0)}{\Gamma(\alpha^q_0(1-\frac{{\kappa}_i}{\kappa_0}))\prod_{j=1}^{{\kappa}_i} \Gamma(\frac{\alpha^q_0}{\kappa_0})}
  \int_{{\rho}_i=0}^1 (1-{\rho}_i)^{\alpha^q_0(1-\frac{{\kappa}_i}{\kappa_0})-1} ({\rho}_i)^{\alpha^q_0\frac{{\kappa}_i}{\kappa_0}-1} ~d{\rho}_i \right)
  ~(\prod_{j=1}^{{\kappa}_i} (\frac{{\pi}_{ij}}{{\rho}_i})^{\frac{\alpha^q_0}{\kappa_0}-1})
  \\
  &=& d(\frac{{\pi}_{i1}}{{\rho}_i},\overset{{\kappa}_i}{\ldots},\frac{{\pi}_{i{\kappa}_i}}{{\rho}_i}\;;~
  \frac{\alpha^q_0}{\kappa_0},\overset{{\kappa}_i}{\ldots},\frac{\alpha^q_0}{\kappa_0})
\end{eqnarray*}
where in the last step we note that the integral (assuming it is well defined) is simply part of the normalisation constant, which we know from the definition of the Dirichlet distribution must be $B(\frac{\alpha^q_0}{\kappa_0},\overset{{\kappa}_i}{\ldots},\frac{\alpha^q_0}{\kappa_0})$.  This gives us
\begin{eqnarray*} (\frac{{\pi}_{i1}}{{\rho}_i},\overset{{\kappa}_i}{\ldots},\frac{{\pi}_{i{\kappa}_i}}{{\rho}_i})
  &\sim& \Dir(\frac{\alpha^q_0}{\kappa_0},\overset{{\kappa}_i}{\ldots},\frac{\alpha^q_0}{\kappa_0})
\end{eqnarray*}

Now that we have all the individual distributions, we can put them together to get the factorised distribution for the case of finite $\kappa_0$.
\begin{eqnarray*}
  {\pi}_{ij} &=& {\rho}_i {\pi}^\prime_{ij}
  \mbox{~~for~} i=1,\ldots,c,~ j=1,\ldots,{\kappa}_i
  \\
  \vf{\rho} &\sim& \Dir({\alpha}^q_1,\overset{c}{\ldots},{\alpha}^q_c)
  \\
  \vf{\pi}^\prime_{i} &\sim& 
  \Dir( \frac{\alpha^q_0}{\kappa_0},\overset{{\kappa}_i}{~\ldots\ldots~},\frac{\alpha^q_0}{\kappa_0} )
  \mbox{~~for~} i=1,\ldots,c
\end{eqnarray*}
Noting that $\lim_{\kappa_0\rightarrow \infty} \frac{\alpha^q_0}{\kappa_0} = \frac{{\alpha}^q_i}{{\kappa}_i}$, we can then take the limit as $\kappa_0$ goes to infinity to get our definition of the weights for the factorised Dirichlet distribution.
\begin{eqnarray*}
  {\pi}_{ij} &=& {\rho}_i {\pi}^\prime_{ij}
  \mbox{~~for~} i=1,\ldots,c,~ j=1,\ldots,\infty
  \\
  \vf{\rho} &\sim& \Dir({\alpha}^q_1,\overset{c}{\ldots},{\alpha}^q_c)
  \\
  \vf{\pi}^\prime_{i} &\sim& \lim_{{\kappa}_i\rightarrow \infty}
  \Dir( \frac{{\alpha}^q_i}{{\kappa}_i},\overset{{\kappa}_i}{\ldots},\frac{{\alpha}^q_i}{{\kappa}_i} )
  \mbox{~~for~} i=1,\ldots,c
\end{eqnarray*}
Thus the weights of a DP can be rewritten as the weights of an equivalent FDP using the above construction.

Putting the vectors and weights together, we get the distribution ${F}_i$ over the weighted vectors in each partition $i$.
\begin{eqnarray*}
  \vf{z}_{ij} &\sim& {G}^q_i \mbox{~~for~} i=1,\ldots,c,~ j=1,\ldots,{\kappa}_i
  \\
  \vf{\pi}^\prime_{i} &\sim& \lim_{{\kappa}_i\rightarrow \infty}
  \Dir( \frac{{\alpha}^q_i}{{\kappa}_i},\overset{{\kappa}_i}{\ldots},\frac{{\alpha}^q_i}{{\kappa}_i} )
  \mbox{~~for~} i=1,\ldots,c
  \\ \text{and thus} \\
  {F}_i &\sim& \DP({G}^q_i,{\alpha}^q_i)
  \mbox{~~for~} i=1,\ldots,c
\end{eqnarray*}
This concludes our proof that, if $F\sim \DP(G^q_0,\alpha^q_0)$, then:
\begin{eqnarray*}
  F &=& \sum_{i=1}^{c} {\rho}_i {F}_i
  \\
  \vf{\rho} &\sim& \Dir({\alpha}^q_1,\overset{c}{\ldots},{\alpha}^q_{c})
  \\
  {F}_i &\sim& \DP({G}^q_i,{\alpha}^q_i)
  \mbox{~~for~} i=1,\ldots,c
\end{eqnarray*}
and thus $\FDP(\vf{G}^q,\vf{\alpha}^q) = \DP(G^q_0,\alpha^q_0)$.

\section{Deriving the KL Divergence}
\label{sec:KLderivation}

The formulation of both the prior and posterior as bounded factorised DPs of the same form simplifies the computation of the KL divergence, because the KL divergence for each respective pair of factors can be computed separately, and then combined.  

First consider the factors for the Dirichlet distributions over the partitions for the different components $i$.  There is a closed-form solution to the KL divergence between two Dirichlet distributions.
\begin{eqnarray*}
  \hspace{2ex}&&\hspace{-7ex}
  \KL\left( \Dir({\alpha}^q_1,\overset{c+1}{\ldots},{\alpha}^q_{c+1})
  \kld \Dir(\alpha^p_0\tfrac{{\alpha}^q_1}{\alpha^q_0},\overset{c+1}{\ldots},\alpha^p_0\tfrac{{\alpha}^q_{c+1}}{\alpha^q_0}) \right)
  \\
  &=& \log\frac{\Gamma(\alpha^q_0)}{\Gamma(\alpha^p_0)}
  +\sum_{i=1}^{c+1}~ \left( -\log\frac{\Gamma({\alpha}^q_i)}{\Gamma(\alpha^p_0\tfrac{{\alpha}^q_i}{\alpha^q_0})} 
  +{\alpha}^q_i(1-\tfrac{\alpha^p_0}{\alpha^q_0}) ( \psi({\alpha}^q_i)-\psi(\alpha^q_0) ) \right)
\end{eqnarray*}
where $\psi$ is the digamma function.

For the bounded DPs for each individual component $i$, there are two factors, a symmetric Dirichlet distribution over the weights and a Gaussian distribution over each vector.  For the symmetric Dirichlet distribution, in the case where ${\kappa}_i = 1$, then the KL for this term is zero, since there is no choice to make for this weight.  In the case where ${\kappa}_i> 1$,
the KL divergence between the posterior and prior versions of these weight distributions again has a closed-form solution.
\begin{eqnarray*}
  \hspace{2ex}&&\hspace{-7ex}
  \KL\left( \Dir(\tfrac{{\alpha}^q_i}{{\kappa}_i},\overset{{\kappa}_i}{\ldots},\tfrac{{\alpha}^q_i}{{\kappa}_i})
  \kld \Dir(\alpha^p_0\tfrac{{\alpha}^q_i}{\alpha^q_0{\kappa}_i},\overset{{\kappa}_i}{\ldots},\alpha^p_0\tfrac{{\alpha}^q_i}{\alpha^q_0{\kappa}_i}) \right)
  \\
  &=& \log\frac{\Gamma({\alpha}^q_i)}{\Gamma(\alpha^p_0\tfrac{{\alpha}^q_i}{\alpha^q_0})}
  -{\kappa}_i \log\frac{\Gamma(\tfrac{{\alpha}^q_i}{{\kappa}_i})}{\Gamma(\alpha^p_0\tfrac{{\alpha}^q_i}{\alpha^q_0{\kappa}_i})}
  +{\alpha}^q_i(1-\tfrac{\alpha^p_0}{\alpha^q_0}) ( \psi(\tfrac{{\alpha}^q_i}{{\kappa}_i})-\psi({\alpha}^q_i) )
\end{eqnarray*}
This term then needs to be summed across components $1\leq i\leq c{+}1$.

For the factors for generating vectors from each individual component of the base distribution, because the different components have been factorised, there is also a closed-form solution to computing these KL divergences.  Each Gaussian component of the posterior's base distribution is compared independently to the Gaussian of the prior's base distribution.
The $\KL$ divergence between two Gaussians (with diagonal covariance with values $\vf{\sigma}$) is:
\begin{eqnarray*}
  \KL({G}^q_i\kld G^p_0)
  &=&
  \tfrac{1}{2} \sum_{h=1}^d ( \frac{({\mu}^q_{ih}-{\mu}^p_{h})^2}{({\sigma}^p_{h})^2}
  +\frac{({\sigma}^q_{ih})^2}{({\sigma}^p_{h})^2} -1
  -\log(\frac{({\sigma}^q_{ih})^2}{({\sigma}^p_{h})^2}) )
  \\
  &=& \tfrac{1}{2} \sum_{h=1}^d ( ({\mu}^q_{ih})^2 + ({\sigma}^q_{ih})^2 - 1
  - \log(({\sigma}^q_{ih})^2) )
\end{eqnarray*}
where the last step assumes $\vf{\mu}^p = \vf{0},~ (\vf{\sigma}^p)^2 = \vf{1}$.
This term then needs to be multiplied by the number $\kappa_i$ of vectors for this component, and summed across components $1\leq i\leq c{+}1$.

Given these exact closed-form solutions for each pair of factors, we can compute the full KL divergence.  We start by combining the formulas for the weight factors, where some terms cancel:
\begin{eqnarray*}
  &&\hspace{-7ex}
  \KL\left( \Dir({\alpha}^q_1,\overset{c+1}{\ldots},{\alpha}^q_{c+1})
  \kld \Dir(\alpha^p_0\tfrac{{\alpha}^q_1}{\alpha^q_0},\overset{c+1}{\ldots},\alpha^p_0\tfrac{{\alpha}^q_{c+1}}{\alpha^q_0}) \right)
  \\&&\hspace{-7ex}
  +\sum_{i=1}^{c+1} \KL\left( \Dir(\tfrac{{\alpha}^q_i}{{\kappa}_i},\overset{{\kappa}_i}{\ldots},\tfrac{{\alpha}^q_i}{{\kappa}_i})
  \kld \Dir(\alpha^p_0\tfrac{{\alpha}^q_i}{\alpha^q_0{\kappa}_i},\overset{{\kappa}_i}{\ldots},\alpha^p_0\tfrac{{\alpha}^q_i}{\alpha^q_0{\kappa}_i}) \right)
  \\
  &=& \log\frac{\Gamma(\alpha^q_0)}{\Gamma(\alpha^p_0)}
  -\sum_{i=1}^{c+1} {\kappa}_i \log\frac{\Gamma(\tfrac{{\alpha}^q_i}{{\kappa}_i})}{\Gamma(\frac{\alpha^p_0{\alpha}^q_i}{\alpha^q_0{\kappa}_i})}
  +(1-\tfrac{\alpha^p_0}{\alpha^q_0}) \sum_{i=1}^{c+1} {\alpha}^q_i ( \psi(\tfrac{{\alpha}^q_i}{{\kappa}_i})-\psi(\alpha^q_0) )
\end{eqnarray*}
Now we can put all these pieces together.
\begin{eqnarray}
  &&\hspace{-7ex} \nonumber
  \KL( \BFDP(\vf{G}^q,\vf{\alpha}^q,\vf{\kappa})
  \kld \BFDP(\vf{G}^p,\vf{\alpha}^q\tfrac{\alpha^p_0}{\alpha^q_0},\vf{\kappa}) )
  \\ \nonumber
  &=& \int_{\vf{\rho}} \int_{\vf{\pi}^\prime} \int_{\vf{v}}
  d(\vf{\rho}\;;~ {\alpha}^q_1,\ldots,{\alpha}^q_{c+1})
  \left( \prod_{i=1}^{c+1} d(\vf{\pi}^\prime_i\;;~ \tfrac{{\alpha}^q_i}{{\kappa}_i},\overset{{\kappa}_i}{\ldots},\tfrac{{\alpha}^q_i}{{\kappa}_i} )
  \prod_{i=1}^{c+1} \prod_{j=1}^{{\kappa}_i} {G}^q_i(\vf{z}_{ij}) \right)
  ~\log\frac{ \BFDP(\vf{G}^p,\vf{\alpha}^q\tfrac{\alpha^p_0}{\alpha^q_0},\vf{\kappa}) }{ \BFDP(\vf{G}^q,\vf{\alpha}^q,\vf{\kappa}) }
  ~ d~{\vf{\rho}}~ d~{\vf{\pi}^\prime}~ d~{\vf{v}}
  \\ \nonumber
  &=&
  \KL\left( \Dir({\alpha}^q_1,\overset{c+1}{\ldots},{\alpha}^q_{c+1})
  \kld \Dir(\alpha^p_0\tfrac{{\alpha}^q_1}{\alpha^q_0},\overset{c+1}{\ldots},\alpha^p_0\tfrac{{\alpha}^q_{c+1}}{\alpha^q_0}) \right)
  \\&& \nonumber
  +\sum_{i=1}^{c+1}~ \KL\left( \Dir(\tfrac{{\alpha}^q_i}{{\kappa}_i},\overset{{\kappa}_i}{\ldots},\tfrac{{\alpha}^q_i}{{\kappa}_i})
  \kld \Dir(\alpha^p_0\tfrac{{\alpha}^q_i}{\alpha^q_0{\kappa}_i},\overset{{\kappa}_i}{\ldots},\alpha^p_0\tfrac{{\alpha}^q_i}{\alpha^q_0{\kappa}_i}) \right)~
  +{\kappa}_i \KL({G}^q_i\kld G^p_0)
  \\ \label{eq:A-KLloss}
  &=& \log\Gamma(\alpha^q_0) -\log\Gamma(\alpha^p_0)
  +\sum_{i=1}^{c+1} {\kappa}_i ( \log\Gamma(\frac{\alpha^p_0{\alpha}^q_i}{\alpha^q_0{\kappa}_i}) -\log\Gamma(\tfrac{{\alpha}^q_i}{{\kappa}_i}) )
  +(\alpha^q_0-\alpha^p_0) \left(
  -\psi(\alpha^q_0) +\sum_{i=1}^{c+1} \tfrac{{\alpha}^q_i}{\alpha^q_0} \psi(\tfrac{{\alpha}^q_i}{{\kappa}_i})
  \right)
  \\&& \nonumber
  +\tfrac{1}{2} \sum_{i=1}^{c+1} {\kappa}_i \sum_{h=1}^d ( \frac{({\mu}^q_{ih}-{\mu}^p_{h})^2}{({\sigma}^p_{h})^2} + \frac{({\sigma}^q_{ih})^2}{({\sigma}^p_{h})^2}
  -\log\frac{({\sigma}^q_{ih})^2}{({\sigma}^p_{h})^2} -1 )
  \\ \nonumber
  &=& \log\Gamma(\alpha^q_0) 
  +\sum_{i=1}^{c+1} {\kappa}_i \left( \log\Gamma(\frac{{\alpha}^q_i}{\alpha^q_0{\kappa}_i}) -\log\Gamma(\frac{{\alpha}^q_i}{{\kappa}_i}) \right)
  +(\alpha^q_0-1) \left(
  -\psi(\alpha^q_0) +\sum_{i=1}^{c+1} \tfrac{{\alpha}^q_i}{\alpha^q_0} \psi(\frac{{\alpha}^q_i}{{\kappa}_i})
  \right)
  \\&& \nonumber
  +\tfrac{1}{2} \sum_{i=1}^{c+1} {\kappa}_i \sum_{h=1}^d ( ({\mu}^q_{ih})^2 + ({\sigma}^q_{ih})^2 -1
  -\log(({\sigma}^q_{ih})^2)  )
\end{eqnarray}
where the last step assumes $\alpha^p_0 = 1,~ \vf{\mu}^p = \vf{0},~ (\vf{\sigma}^p)^2 = \vf{1}$.
Equation~\eqref{eq:A-KLloss} gives the KL portion of the loss when we are given the full set $\vf{\kappa}$ of numbers of vectors ${\kappa}_i$ generated for each component $i$.

Equation~\eqref{eq:A-KLloss} is approximately linear in ${\kappa}_i$, when the variation in ${\kappa}_i$ is fairly small relative to the values of ${\kappa}_i$.  The Gaussian term is exactly linear in ${\kappa}_i$, and the terms $\psi(\frac{{\alpha}^q_i}{{\kappa}_i})$ and ${\kappa}_i \left( \log\Gamma(\frac{{\alpha}^q_i}{\alpha^q_0{\kappa}_i}) -\log\Gamma(\frac{{\alpha}^q_i}{{\kappa}_i}) \right)$ are both approximately linear in ${\kappa}_i$.  This allows us to approximate the expectation over ${\kappa}_i$ of this loss as this loss of the expectation over ${\kappa}_i$, as discussed in Section~\ref{sec:KL}.
In this case, this approximation of the full KL divergence is: 
\begin{eqnarray}
  &&\hspace{-7ex} \nonumber
  \KL( \BFDP(\vf{G}^q,\vf{\alpha}^q,\vf{\kappa})
  \kld \BFDP(\vf{G}^p,\vf{\alpha}^q\tfrac{\alpha^p_0}{\alpha^q_0},\vf{\kappa}) )
  \\ \label{eq:A-KLEloss}
  &\approx&
  \log\Gamma(\alpha^q_0)
  +(\alpha^q_0 -1) \left(
  \psi(\tfrac{\alpha^q_0}{\kappa_0})
  -\psi(\alpha^q_0)   \right)
  +\kappa_0 \left(
  \log\Gamma(\frac{1}{\kappa_0}) -\log\Gamma(\frac{\alpha^q_0}{\kappa_0})  \right)
  \\&& \nonumber
  +\tfrac{1}{2}\kappa_0 \sum_{i=1}^{c+1} \frac{{\alpha}^q_i}{\alpha^q_0}
   \sum_{h=1}^d \left( ({\mu}^q_{ih})^2 + ({\sigma}^q_{ih})^2 - 1 - \log(({\sigma}^q_{ih})^2) \right)
\end{eqnarray}

In the case where ${\kappa}_i=1$ for all $1\leq i\leq c+1$, then each Gaussian ${G}^q_i$ is only used once, and there is no Dirichlet KL term for the individual components.  This gives us a full KL divergence of:
\begin{eqnarray}
  &&\hspace{-7ex} \nonumber
  \KL( \BFDP(\vf{G}^q,\vf{\alpha}^q,\vf{1})
  \kld \BFDP(\vf{G}^p,\vf{\alpha}^q\tfrac{\alpha^p_0}{\alpha^q_0},\vf{1}) )
  \\ \label{A-Onesample-KLloss}
  &=& \log\Gamma(\alpha^q_0) +\sum_{i=1}^{c+1} \left( 
  \log\frac{\Gamma(\tfrac{{\alpha}^q_i}{\alpha^q_0})}{\Gamma({\alpha}^q_i)}
  +{\alpha}^q_i(1-\tfrac{1}{\alpha^q_0}) ( \psi({\alpha}^q_i)-\psi(\alpha^q_0) )
  \right)
  \\&& \nonumber
  +\tfrac{1}{2}\sum_{i=1}^{c+1} \sum_{h=1}^d \left( ({\mu}^q_{ih})^2 + ({\sigma}^q_{ih})^2 - 1 - \log(({\sigma}^q_{ih})^2) \right)
\end{eqnarray}

\end{document}